%% file: sample-sigconf.tex
\documentclass[sigconf]{acmart}

\AtBeginDocument{%
  }

\usepackage{graphicx}
\usepackage{enumitem}
\usepackage{multirow}
\usepackage{placeins}
\usepackage{array}
\usepackage{xcolor}
\usepackage{bm}
\usepackage{colortbl}
\usepackage{threeparttable}
\usepackage{afterpage}
\usepackage{tablefootnote}
\usepackage{tikz}
\usepackage{pgfplots}
\pgfplotsset{compat=1.18} % adjust if needed

\usepackage{pifont}

\newcolumntype{B}{>{\columncolor{blue!10}\bfseries\color{blue}}c}

\definecolor{lightblue}{rgb}{0.85, 0.95, 1}
\definecolor{lightorange}{rgb}{1, 0.95, 0.85}
\definecolor{lightpink}{rgb}{1, 0.9, 0.95}
\definecolor{lightgreen}{rgb}{0.56, 0.93, 0.56}

\definecolor{pastelblue}{rgb}{0.68, 0.85, 0.90}
\definecolor{darkergreen}{rgb}{0.00, 0.5, 0.00}

\definecolor{color1}{RGB}{120,159,124}
\definecolor{color2}{RGB}{199,115,100}
\definecolor{color3}{RGB}{252,104,58}
\definecolor{color4}{RGB}{250,168,68}
\definecolor{color5}{RGB}{121,137,184}
\definecolor{color6}{RGB}{167,98,236}

\usepackage[absolute,overlay]{textpos}

\copyrightyear{2026}
\acmYear{2026}
\setcopyright{cc}
\setcctype{by} % <-- CC-BY (Option a)
\acmConference[WWW '26]{Proceedings of the ACM Web Conference 2026}{April 13--17, 2026}{Dubai, United Arab Emirates}
\acmBooktitle{Proceedings of the ACM Web Conference 2026 (WWW '26), April 13--17, 2026, Dubai, United Arab Emirates}
\acmISBN{979-8-4007-2307-0/2026/04}
\acmDOI{10.1145/3774904.3792509}

\AtBeginDocument{%
  \setlength{\textfloatsep}{6pt} % 顶/底部浮动体与正文距离（关键）
  \setlength{\intextsep}{6pt}    % 文中浮动体与正文距离
  \setlength{\floatsep}{6pt}     % 浮动体之间距离
}
\begin{document}

% =========================
% Title
% =========================
% \title[DeepSVU]{DeepSVU: Towards In-depth Security-oriented Video Understanding via Unified Physical-world Regularized MoE}
\title[DeepSVU]{DeepSVU: Towards In-depth Security-oriented Video Understanding via Unified Physical-world Regularized MoE}

% =========================
% Authors (must match rightsreview exactly)
% =========================
\author{Yujie Jin}
\email{yjjin0727@stu.suda.edu.cn}
\affiliation{%
  \institution{School of Computer Science and Technology, Soochow University}
  \city{Suzhou}
  \country{China}
}

\author{Wenxin Zhang}
\email{wxzhang0530@stu.suda.edu.cn}
\affiliation{%
  \institution{School of Computer Science and Technology, Soochow University}
  \city{Suzhou}
  \country{China}
}

\author{Jingjing Wang}
\authornote{Corresponding Author: Jingjing Wang.}
\email{djingwang@suda.edu.cn}
\affiliation{%
  \institution{School of Computer Science and Technology, Soochow University}
  \city{Suzhou}
  \country{China}
}

\author{Guodong Zhou}
\email{gdzhou@suda.edu.cn}
\affiliation{%
  \institution{School of Computer Science and Technology, Soochow University}
  \city{Suzhou}
  \country{China}
}

% Short author list for running headers
\renewcommand{\shortauthors}{Yujie Jin, Jingjing Wang, Wenxin Zhang \& Guodong Zhou}

% =========================
% Abstract
% =========================
\input{contents/abstract}

% =========================
% CCS Concepts
% =========================

% =========================
% Keywords
% =========================
% \keywords{Security-oriented Video; Physical-world Modeling; Video-LLMs; Regularized Mixture-of-Experts (MoE)}
% =========================
% Teaser figure (optional)
% =========================
\begin{teaserfigure}
\vspace{-1.5em} % 图整体往上提（可调：-0.6em 到 -2.0em）
  \includegraphics[width=\textwidth]{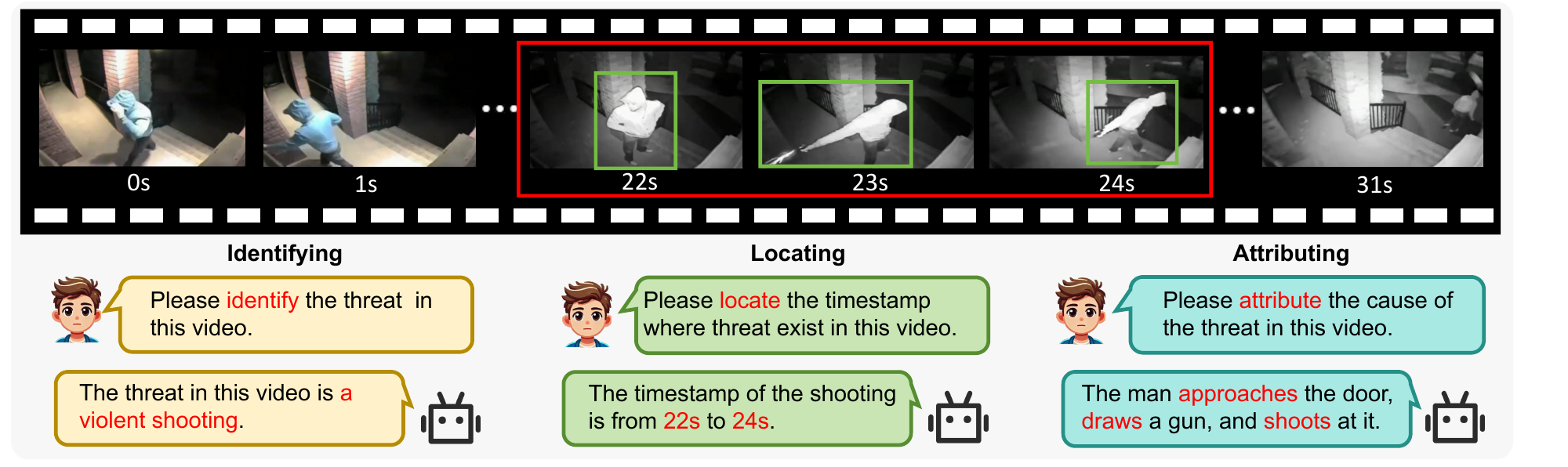}
  \setlength{\abovecaptionskip}{-2ex}
  \caption{An example illustrating the DeepSVU task and the output of the proposed UPRM approach. In a surveillance video, UPRM can not only precisely identify and locate threatening segments but also attribute their causes.}
  \Description{Overview figure for the DeepSVU task and UPRM outputs.}
  \label{fig:intro}
\end{teaserfigure}

\maketitle

% =========================
% Main content
% =========================
\input{contents/introduction}
\input{contents/related_work}

\input{contents/approach}
\input{contents/experiments}

\input{contents/analysis}

\input{contents/conclusion}

\begin{acks}
This work was supported by two NSFC grants, i.e., No.62576234, No.62376178 and sponsored by CIPS-LMG Huawei Innovation Fund. This work was also supported by Collaborative Innovation Center of Novel Software Technology and Industrialization, and a Project funded by the Priority Academic Program Development of Jiangsu Higher Education Institutions (PAPD).
\end{acks}

% =========================
% Acknowledgments (if any)
% =========================
% \begin{acks}
% ...
% \end{acks}

% =========================
% Bibliography
% =========================
\bibliographystyle{ACM-Reference-Format}


% =========================
% Appendix (if any, and only within the allowed extra pages)
% =========================
% \appendix

\input{contents/supplementary}

\end{document}

%% file: contents/abstract.tex
\begin{abstract}
In the literature, prior research on Security-oriented Video Understanding (SVU) has predominantly focused on detecting and locating the threats (e.g., shootings, robberies) in videos, while largely lacking the effective capability to generate and evaluate the threat causes.
Motivated by these gaps, this paper introduces a new chat paradigm SVU task, i.e., In-depth Security-oriented Video Understanding (DeepSVU), which aims to not only identify and locate the threats but also attribute and evaluate the causes of threatening segments in detail. Furthermore, this paper reveals two key challenges in the proposed task: 1) how to effectively model the coarse-to-fine physical-world information (e.g., human behavior, object interactions and background context) to boost the DeepSVU task, and 2) how to adaptively trade off these factors. 
% Addressing these challenges is crucial for improving VAD, especially for identifying, locating, and attributing anomalies. 
To tackle these challenges, this paper proposes a new Unified Physical-world Regularized MoE
 (UPRM) approach.
Specifically, UPRM incorporates two key components: the Unified Physical-world Enhanced MoE (UPE) Block and the Physical-world Trade-off Regularizer (PTR), to address the above two challenges, respectively.
Extensive experiments conduct on our DeepSVU instructions datasets
(i.e., UCF-C instructions and CUVA instructions) demonstrate that UPRM outperforms several advanced Video-LLMs as well as non-LLM approaches.
such information.These justify the importance of the coarse-to-fine physical-world information  in the DeepSVU task and demonstrate the effectiveness of our UPRM in capturing such information.
\end{abstract}

%% file: contents/introduction.tex
\section{Introduction}
% \vspace{-1 ex}
% \begin{figure*} 
%     \includegraphics[width=0.49\textwidth]{figure/piechart1.pdf}
%     \includegraphics[width=0.49\textwidth]{figure/piechart2.pdf}
     
%  \caption{The statistics of physical-world information for (a) the UCF-C 
%  instructions dataset and (b) the CUVA  instructions. These pie charts illustrate the imbalanced distribution of these information, with coarse-grained information and human-pose comprising the largest and second-largest segments. The surrounding images provide examples corresponding to each fine-grained physical-world information. The proportions are computed based on the frequency of each physical-world descriptions in the dataset, with segment size reflecting its relative occurrence.}
%   \label{fig:dataset}
% \end{figure*}

% In recent years, Security-oriented Video Understanding (SVU) has attracted growing attention due to its extensive applications in intelligent video surveillance, content moderation, and environmental monitoring.
% In recent years, Security-oriented Video Understanding (SVU) has attracted extensive attention, focusing on task variations such as video harmful content detection\cite{yousaf2022deep,edstedt2022vidharm,jinyujiehawkeye}, and UAV or surveillance-based anomaly detection\cite{jin2022anomaly,bozcan2020air}.
Security-oriented Video Understanding (SVU), generally involving different task variations such as video harmful content detection \cite{yousaf2022deep,edstedt2022vidharm,jinyujiehawkeye} and UAV or surveillance-based anomaly detection \cite{jin2022anomaly,zhou2023dual,vadclip}, mainly aims to automatically recognize potential security-related threats (e.g., shootings, robberies) in videos, for advancing multimedia research with significant implications for intelligent monitoring and public safety applications.
Previous SVU approaches \cite{hasan2016learning,Tran_Bourdev_Fergus_Torresani_Paluri_2015,sun2023hierarchical} predominantly rely on non-Large Language Model (non-LLM) approaches. Recent advances in Video-LLMs \cite{lin2023video,videochat} have drawn increasing attention, achieving prominent results in various video-related tasks. Thus, in the field of SVU, some studies have also started using Video-LLMs to address threat-related tasks, such as Holmes \cite{zhang2024holmes}, Ex-VAD \cite{huang2025exvad} and Hawkeye \cite{jinyujiehawkeye}. 
Nevertheless, these works primarily focus on identifying and locating threats, while lacking effective capability to generate and evaluate the threat causes, and failing to model the detailed semantics of the physical world in videos.

Building on these foundations, this paper proposes a new, In-depth Security-oriented Video Understanding (DeepSVU) task
, which not only identifies and locates threats but also attributes and detailly evaluates the causes of  threatening segments in detail via Video-LLM.
For example, as illustrated in Figure  \ref{fig:intro}, a threat occurs between 22 and 24 seconds, where a man near a door is seen shooting at it. This new task can locate the exact timestamps and explain the root causes of threats in a surveillance video.  It transforms traditional threat detection from simply ``\textit{identifying problems}'' to comprehensively ``\textit{understanding problems}''.
Since this task can not only  precisely identify threats, but also locate their timestamps and provide detailed explanations of their causes, it could make threat detection systems more efficient and precise,  contributing to the development of a comprehensive and intelligent video threat monitoring and tracking system. Thus, this makes the task highly worthwhile. Despite this, we believe this new task faces at least two primary challenges, which are outlined as follows.

On one hand, how to model coarse-to-fine physical-world information to boost the new DeepSVU task  is challenging. Existing Video-LLMs \cite{pandagpt,li20243dmit} generally focus on coarse-grained video information (i.e., the general video representations modeling videos in a broad manner), while often overlooking fine-grained physical-world information (e.g., human behavior, object interactions, and background context) which potentially contributes to the new task. As illustrated in Figure \ref{fig:intro}, incorporating physical-world information such as human poses (e.g., shooting), object relationships (e.g., a man near a door), and visual background features (e.g., a house) can significantly enhance the detection of threatening segments. 
However, the heterogeneity inherent in fine-grained physical-world information—arising from diverse model architectures and encoders—poses significant challenges. A single, fixed-capacity transformer-based model often lacks the ability to effectively capture and integrate these crucial details, hindering Video-LLMs from achieving a nuanced understanding of scenes.
Recently, the Mixture of Experts (MoE) \cite{onellm,li2024cumo,luo2025omnisila} paradigm has shown remarkable scalability in combining diverse representations across multiple modalities. Motivated by this progress, a well-designed approach to DeepSVU could adopt MoE architecture to capture fine-grained physical-world details such as human pose and object relations, as well as coarse-grained visual information, thereby enhancing the nuanced understanding capabilities of Video-LLMs.

On the other hand, how to effectively trade off the unified physical-world information presents another challenge. A straightforward approach is to adopt a basic MoE to assign corresponding weights to the various physical-world information. However, as shown in Figure \ref{fig:dataset},  we collect statistical physical-world information from two datasets: CUVA instructions and UCF-C instructions (see Section \ref{first:fourth_chapter}). It can be observed that coarse-grained and human-pose information dominate the physical-world data (i.e., coarse-grained at 100\%, and human-pose at 41\% and 46\%), indicating an imbalance in the distribution of coarse-to-fine information. This imbalance may cause the basic MoE model to be biased towards the more frequently occurring information, potentially leading to misclassification of threatening video segments as non-threatening. 
For example, Figure \ref{fig:bar 7} illustrates that coarse-grained expert tends to be more extensively trained, followed by the pose expert, often resulting in them having the highest weights.
Motivated by this, a better-designed approach to DeepSVU should trade off the coarse-to-fine physical-world information, thereby alleviating the issue of data imbalance.

To tackle the aforementioned challenges in the DeepSVU task, this paper proposes an Unified Physical-world Regularized MoE (UPRM) approach.
 Specifically, to resolve the first challenge, we design a Unified Physical-world Enhanced MoE (UPE) block. 
  This block comprises three fine-grained experts and one coarse-grained expert to comprehensively model physical-world features at varying levels of granularity. 
 To address the second challenge, we design a Coarse-to-Fine Physical-world Trade-off Regularizer (PTR) that mitigates biases between coarse-grained and fine-grained physical-world representations via a Gated Physical-world Trade-off Loss (GTL). Especially, we construct two DeepSVU instructions datasets (i.e., UCF-C instructions and CUVA instructions) based on two  available datasets (i.e., UCF-Crime \cite{ucf-crime}, CUVA \cite{cuva}) to evaluate the effectiveness of UPRM.  Extensive experiments reveal that UPRM outperforms several advanced Video-LLMs and non-LLM approaches.

%  Overall, our main contributions are summarized as follows:
 
% \textbullet ~We propose a new chat paradigm VAD task, i.e., \textbf{Inter}actable Video \textbf{A}nomaly \textbf{i}dentifying, \textbf{l}ocating, and \textbf{a}ttributing (DeepSVU), which could not only identify and locate anomalies but also attribute their causes, enabling VAD systems to be more precise in monitoring.

% \textbullet ~We propose a new \textbf{L}LM\textbf{-}driven \textbf{C}oarse-to-Fine \textbf{P}hysical-\textbf{W}orld Modeling (UPRM) approach, which models not only  coarse-grained video representation but also fine-grained physical-world information to the new DeepSVU task.
% Specifically, our UPRM approach designs a Coarse-to-Fine \textbf{P}hysical World \textbf{E}nhanced \textbf{M}oE (PEM) block and a \textbf{P}hysical World \textbf{T}rade-off \textbf{R}egularizer (PTR), effectively addressing the challenges of physical-world modeling and trade-off, respectively helping the model to further improve its ability in precisely identifying, locating, and attributing of anomalies.

% \textbullet ~We conduct extensive experiments on our constructed DeepSVU instructions datasets, and the experimental results demonstrate that the proposed UPRM approach surpasses several advanced Video-LLMs and traditional non-LLM approaches. This justifies the importance of the fine-grained physical-world information and the effectiveness of the proposed UPRM approach in capturing such information.

 \begin{figure*}
    \centering
    \includegraphics[scale=0.52]{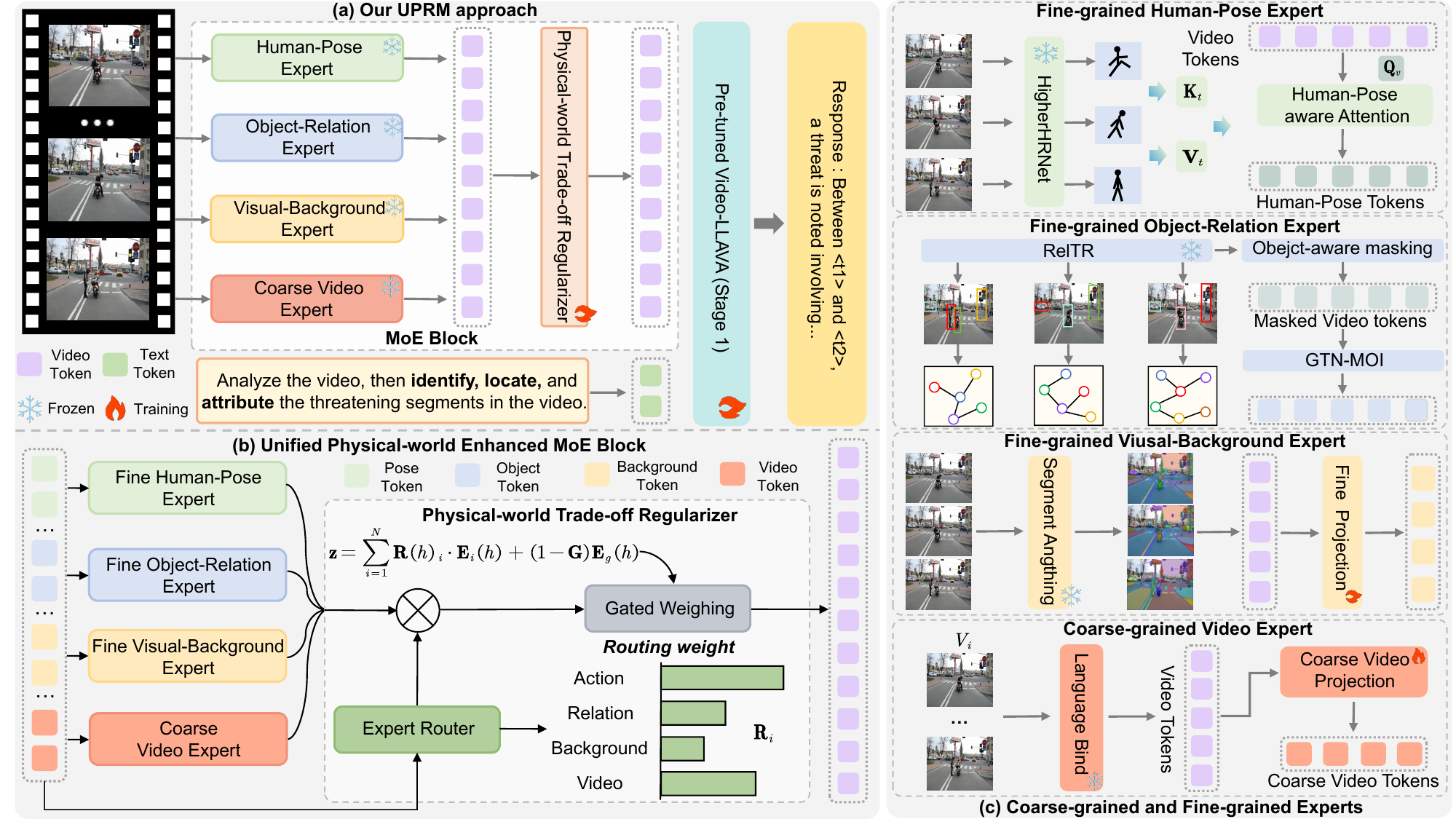}
    \setlength{\abovecaptionskip}{0.9 ex}
    \setlength{\belowcaptionskip}{-1 ex}
        \caption{(a) The overall architecture of the Unified Physical-world Regularized MoE (UPRM) approach. (b) Unified Physical-world Enhanced MoE Block. It consists of two main components: Coarse-to-Fine Experts and Physical-world Trade-off Regularizer. (c) Coarse-grained and Fine-grained Experts, which is used to model the physical-world information.}
    \label{fig:model}
\end{figure*}

%% file: contents/related_work.tex
% 22222

\section{Related Work}
 
\subsection{Security-oriented Video Understanding}
Security-oriented Video Understanding (SVU), generally involving different task variations, such as video harmful content detection \cite{yousaf2022deep,jinyujiehawkeye} and anomaly detection in UAV or surveillance scenarios \cite{ma2025sherlock,vadclip,luo2025omnisila}), mainly aims to detect and locate potential security-related risks (e.g., shootings, robberies) in videos. SVU methods fall into two categories: non-LLM and LLM-based approaches. 
Non-LLM methods employ diverse strategies: Sultani et al.\cite{ucf-crime} propose multi-instance learning for video segment analysis; Zhou et al. \cite{zhou2023dual} integrate global-local transformer modules for temporal modeling, while VadCLIP \cite{vadclip} combines visual-language alignment with classification. 
LLM-based approaches such as Hawkeye \cite{jinyujiehawkeye}, Holmes \cite{zhang2024holmes}, and Ex-VAD \cite{huang2025exvad} attempt threat detection but primarily focus on identifying and locating threats, while lacking effective capability to generate and evaluate the threat causes, and failing to model the detailed semantics of the physical world.
Unlike the aforementioned studies, we propose a new DeepSVU task, which not only identifies and locates threats but also attributes and evaluates the causes of threatening segments in detail, enabling systems to be more precise and intelligent in threat monitoring.  

\subsection{Video-oriented Large Language Models}

Video-LLMs have advanced video analysis by broadly falling into four categories: LLM‐based video agents, pre-trained models, instruction fine-tuned models, and hybrid approaches \cite{tang2023video}. LLM-based agents integrate GPT-4 with auxiliary modules for video processing \cite{zhang2024mm}, pre-trained models unify multimodal data via fusion techniques \cite{chen2024vast}, instruction-tuned systems employ lightweight adapters (e.g., Video-LLaMA \cite{zhang2023video}) for efficient alignment, and hybrid approaches combine instruction tuning with agent frameworks \cite{videochat,munasinghe2023pg}. 
Despite their robust capabilities in video understanding, these Video-LLMs face limitations in analyzing threat and cannot effectively trade off different physical-world information. Different from existing Video-LLMs, we propose a new approach that not only identifies video threats but also attributes their causes. This approach incorporates the Unified Physical-world Enhanced MoE (UPE) and the
Physical-world Trade-off Regularizer (PTR), effectively capturing coarse-to-fine physical-world information and weighing their contributions.

%% file: contents/approach.tex
% 33333

% \begin{figure}
%     \centering
%     \hspace{-0.022\textwidth}
%     \includegraphics[width=0.5\textwidth]{figure/static.png}
%     \setlength{\abovecaptionskip}{-1ex}  % Decrease this value to bring the caption closer to the image
%     \setlength{\belowcaptionskip}{-3ex}
%  \caption{Statistics of our Aila dataset.}
%   \label{fig:dataset}
% \end{figure}

\section{Approach}
\noindent \textbf{Challenges.} The DeepSVU task faces two key challenges: physical-world            modeling and trade-off. To address these challenges, we propose a Unified Physical-world Enhanced MoE (UPE) Block for detailed modeling and the Physical-world Trade-off Regularizer (PWR) to mitigate biases between coarse-grained and fine-grained physical-world representations.
Moreover, we select Video-LLaVA \cite{lin2023video} as the backbone for our UPRM approach.

\noindent \textbf{Task Definition.} Given an video $V$ consisting of $T$ segments, each segment is labeled as 1 or 0, where 1 indicates the presence of a threat  and 0 indicates its absence, along with a timestamp $(s,e)$ and cause $c$.
The goal of DeepSVU is to interactively identify, locate, and attribute threatening segments in $V$. It generates a set of segments $\{(s_1, e_1, c_1),...,(s_i, e_i, c_i),...,(s_n, e_n, c_n)\}$, where $s_i$ and $e_i$ denote the start and end times of the $i$-th threatening segment, and $c_i$ represents its causal attribution.

\subsection{Unified Physical-world Enhanced MoE Block }
\label{challenge1}
As shown in Figure \ref{fig:model}, we design a Unified Physical-world Enhanced Mixture-of-Experts (UPE) Block to address the coarse-to-fine physical-world modeling challenge. Specifically, we design three fine-grained physical-world experts and one coarse-grained video expert to model physical-world features. In the following, we will provide a detailed introduction to the three experts respectively.

% 第一个模块
% 11111111111111111
% 1111111111111

\textbf{Fine-grained Human-Pose Expert (HPE).} HPE is designed to model fine-grained human poses. In HPE, given a video sequence, we first employ HigherHRNet \cite{cheng2020higherhrnet} to compute a pose vector $\mathbf{x^t_i}$, encoding the coordinates of 17 human joint positions for each individual in a frame, where $i$ denotes the index of the frame.
Consequently, the complete set of pose vectors, $\mathbf{{T}_{\mathit{a}}} = \{\mathbf{x}^{\mathit{t}}_{\mathit{1}}, \dots, \mathbf{x}^{\mathit{t}}_{\mathit{i}}, \dots,\mathbf{x}^{\mathit{t}}_{\mathit{n}}\}$ 
represents the entire video sequence, where $n$ denotes the total number of frames.

Further, since the variability in motion patterns, occlusions, and the subtle interactions between different joints make it difficult to model poses precisely using raw pose data alone, effectively capturing the dynamics and intricacies of human poses from these pose vectors presents significant issues. To address such an issue, we design a Human-Pose aware Attention Mechanism inspired by \cite{mao2021multi,zhong2023attt2m,jinyujiehawkeye}, which leverages graph-based representations to model relationships between different joints, allowing the system to focus on most relevant parts of the pose data. By constructing a graph where nodes represent joint positions and edges capture their interactions, the attention mechanism can effectively enhance the video vectors, leading to more robust and precise pose recognition.
 
More specifically, the Human-Pose aware Attention Mechanism is designed to enhance the video vectors $\mathbf{T}_{\mathit{v}} = \{\mathbf{x}^{\mathit{v}}_1, \dots, \mathbf{x}^{\mathit{v}}_{\mathit{i}}, \dots, \mathbf{x}^{\mathit{v}}_n\}
$, which are generated by the video encoder, using $\mathbf{{T}_{\mathit{a}}}$. For each node $g_k$ in $\mathbf{x}^{\mathit{t}}_{\mathit{i}}$, attention weights $\beta_{kl}$ with its adjacent node $g_l$ are computed as follows: 
\setlength{\abovedisplayskip}{-0.1pt} 
\setlength{\belowdisplayskip}{-0.1pt} 
\begin{equation}
    \beta_{kl} = \text{softmax}\left( \frac{(\mathbf{V}g_k) \cdot (\mathbf{V}g_l)}{\sqrt{p}} \right)
\end{equation}
where $\mathbf{V}$ denotes the weight matrix, and $g_k$ and $g_l$ represent the features of nodes $g_k$ and $g_l$, respectively, with $p$ as the feature dimension.
The weighted features $\hat{g}_k$ for node $g_k$ are then aggregated using $\beta_{kl}$: $\hat{g}_k = \sum_{l \in \mathcal{M}(g_k)} \beta_{kl} \cdot g_l$, where $\mathcal{M}(g_k)$ indicates the neighborhood of node $g_k$. The updated feature at node $g_k$ is given by: \(g_k' = \text{ReLU}(\mathbf{V}[\hat{g}_k, g_k])\). $[\hat{g}_k, g_k]$ represents the concatenation of $\hat{g}_k$ and $g_k$.

After applying the attention mechanism, we enhance $\mathbf{T}_{\mathit{v}}$  using a cross-attention strategy. The computation begins with the query $\mathbf{Q}_{\mathit{v}}$, the key $\mathbf{K}_{\mathit{t}}$, and the value $\mathbf{V}_{\mathit{t}}$. The initial output of the cross-attention layer is defined as $\mathbf{{T}_{\mathit{c}}} = \mathbf{V}_{\mathit{t}} \operatorname{softmax}\left(\mathbf{{Q}_{\mathit{v}}}^\top \mathbf{{K}_{\mathit{t}}}\right)$. We integrate each sub-layer into a residual structure, followed by layer normalization (LN) \cite{layernorm}. The operations are as follows:

%  \begin{equation}
% \begin{split}
% \mathbf{T}'_c &= \mathrm{LN}(\mathbf{T}_c + \mathrm{MSA}(\mathbf{T}_c)) \\
% \mathbf{T}_{c+1} &= \mathrm{LN}(\mathbf{T}'_c + \mathrm{FFN}(\mathbf{T}'_c))
% \end{split}
% \end{equation}
  
% \setlength{\belowdisplayskip}{-0.1pt} 
\begingroup
\setlength{\abovedisplayskip}{-7pt}
\setlength{\belowdisplayskip}{1pt} 

\begin{gather}
\mathbf{T}_c' = \mathrm{LN}\bigl( \mathbf{T}_c + \mathrm{MSA}(\mathbf{T}_c) \bigr) \label{eq:trans1} \\
\mathbf{T}_{c+1}' = \mathrm{LN}\bigl( \mathbf{T}_c' + \mathrm{FFN}(\mathbf{T}_c') \bigr) \label{eq:trans2}
\end{gather}
\endgroup
where $\mathrm{LN}(\cdot)$ denotes layer normalization, $\mathrm{MSA}(\cdot)$ refers to the multi-head self-attention layer, and $\mathrm{FFN}(\cdot)$ denotes the Feed-Forward Network. Following these transformations, the enriched pose-aware video tokens are represented by $\mathbf{{T}}_{\mathit{c}+1}$.

% 第2个模块
%22222222222222
% 22222222222222222

\textbf{Fine-grained Object-Relation Expert (ORE)}. ORE is designed to model fine-grained object relations. As depicted in ORE, the structure is designed to effectively delineate the dynamics of object relationships. We first employ the Relation Transformer (RelTR) \cite{reltr}, a pioneering one-stage methodology for constructing object-relation graphs, to facilitate the creation of a series of interconnected graphs $G=\{G_1,\ldots,G_i,\ldots,G_n\}$ across a continuum of video frames. Each graph $G_i =(V_i, \mathcal{E}_i)$ encapsulates the object-relation network for frame $i$, with $V_i = \{(v_{i,1}, d_{i,1}), \ldots, (v_{i,k}, d_{i,k})\}$ representing a collection of $k$ identified entities, tagged by category $v$ and associated bounding dimensions $d$. The connection set $\mathcal{E}_i$ comprises directed linkages $\{v_{i,p}, s_{i,(p,q)}, v_{i,q}\}$, where each $s_{i,(p,q)}$ is a specific interaction descriptor. 

Further, despite the efficient one-stage approach provided by Relation Transformer (RelTR) for constructing object relationship graphs, it still faces several issues in capturing the complex and dynamic relationships between objects \cite{reltr}. For example, factors such as dynamic interactions between objects, partial occlusions, and diverse environmental conditions can all affect the accuracy and completeness of the relationship graph. These limitations make it difficult to fully and precisely model the fine-grained relationships between objects using traditional relationship graphs, thereby impacting the effectiveness of pose recognition and physical-world understanding.

In this way, we design a Graph Transformer Network for Masked Object Interactions (GTN-MOI) to enhance the usage of relational data derived from objects by following \cite{qu2023layoutllm,jinyujiehawkeye}. We begin by applying precise object masking, which selectively conceals non-critical portions of each object based on their bounding boxes, yielding a set of obscured frame representations $\mathbf{y}^{\mathit{v}}_\mathit{i} \in \mathbf{Y}_{\mathit{v}}$. Consequently, this process generates a series of obfuscated video tokens $\mathbf{Y}_{\mathit{m}} = \{\mathbf{y}^{\mathit{m}}_1, \dots, \mathbf{y}^{\mathit{m}}_i, \dots, \mathbf{y}^{\mathit{m}}_n\}
$.
To augment the representation of each regional component within frames, GTN-MOI employs a cascade of graph transformer layers (GTL) \cite{graphtransformer} that enhance local neighborhood information. We utilize $L$ GTL layers, and denote $\mathbf{F^\ell}$ as the feature matrix at the $\ell$-th layer. The progression through these layers involves a two-part transformation as follows:
\begin{gather}
\mathbf{Z^{(\ell)}} = \mathbf{D}^{-\frac{1}{2}} \mathbf{C} \mathbf{D}^{-\frac{1}{2}} \mathbf{F^{(\ell)}} \\
\mathbf{F}^{(\ell+1)} = \sigma(\mathbf{Z}^{(\ell)} \mathbf{W}^{(\ell)})
\end{gather}
where $\sigma$ represents a nonlinear activation function, $\mathbf{C}$ is the adjacency matrix reflecting the connections within the graph $G_i$, $\mathbf{D}$ is the degree matrix corresponding to $\mathbf{C}$, with $\mathbf{D}{ii}=\sum_j \mathbf{C}{ij}$ indicating the sum of connections for each node, and $\mathbf{W}^{(\ell)}$ is the trainable matrix specific to each layer. After the transformations are applied through all layers, the enhanced region-specific video tokens are synthesized by a Feed-Forward Network (FFN), culminating in the final masked video representation $\mathbf{{Y}_{\mathit{s}}} = \mathrm{FFN}(\mathbf{F}^{(\ell)})$.

% 第三个模块
% 333333
% 333333

\textbf{Fine-grained Visual-Background Expert (VBE)}. VBE is designed to model fine-grained visual backgrounds. The VBE meticulously extracts and processes visual-background information from video sequences. We incorporate SAM\cite{kirillov2023segment}, a powerful model for segmenting and identifying various elements within a scene, to facilitate this process.
For each frame \( i \) in a video \( V \), the frame is fed into SAM as \( \mathbf{y}^{\mathit{t}}_i = \text{SAM}(V_i) \), where the token \( \mathbf{y}^{\mathit{t}}_i \) encapsulates the visual background information. We aggregate these tokens over the entire video to form the collection $\mathbf{{T}_{\mathit{b}}} = \{\mathbf{y}^{\mathit{t}}_1, \dots, \mathbf{y}^{\mathit{t}}_i, \dots, \mathbf{y}^{\mathit{t}}_n\}$, where $n$ is the total number of frames.

Since SAM's exceptional capability in capturing detailed visual background features \cite{sam,zhang2023survey}, there is no need to employ additional complex methods for feature extraction enhancement. Instead, we utilize a series of Feed-Forward Networks (FFNs) to refine these visual-background tokens. The transformation applied to each background token $\mathbf{y}^{\mathit{t}}_i$ to enhance its representation is defined by the equation: $\mathbf{y}^{\mathit{t'}}_i = \text{FFN}(\mathbf{y}^{\mathit{t}}_i)$

In this study, we design a lightweight FFN network to ensure that essential physical-world features are effectively highlighted and consistently maintained throughout the video sequence. This enhancement bolsters the representational capabilities of the visual-background tokens $\mathbf{y}^{\mathit{t'}}_i$, setting the stage for further processing within the model pipeline.

\textbf{Coarse-grained Video Expert (CVE)}. The CVE is specifically designed to extract coarse-grained physical-world information from video sequences. In this work, we employ LanguageBinds within Video-LLaVA, which leverages the ViT-L14 architecture from CLIP. Specifically, for each frame \( i \) in a video \( V \), the frame representation is computed as \( \mathbf{T}_{\mathit{g}} = \text{FFN}(\text{LanguageBinds}(V_i)) \). This process generates coarse-grained tokens for the entire video, denoted as \( \mathbf{T}_{\mathit{g}} = \{ \mathbf{t}_1^{\mathit{g}}, \dots, \mathbf{t}_i^{\mathit{g}}, \dots, \mathbf{t}_n^{\mathit{g}} \} \), where \( n \) is the number of frames in \( V \).

After designing the three fine-grained and one coarse-grained expert, we propose a Physical-world Trade-off Regularizer to adaptively trade off among the four experts, detailed as follows.

\begin{figure} % 这里改为单栏浮动体
    \centering
    \includegraphics[width=1\linewidth]{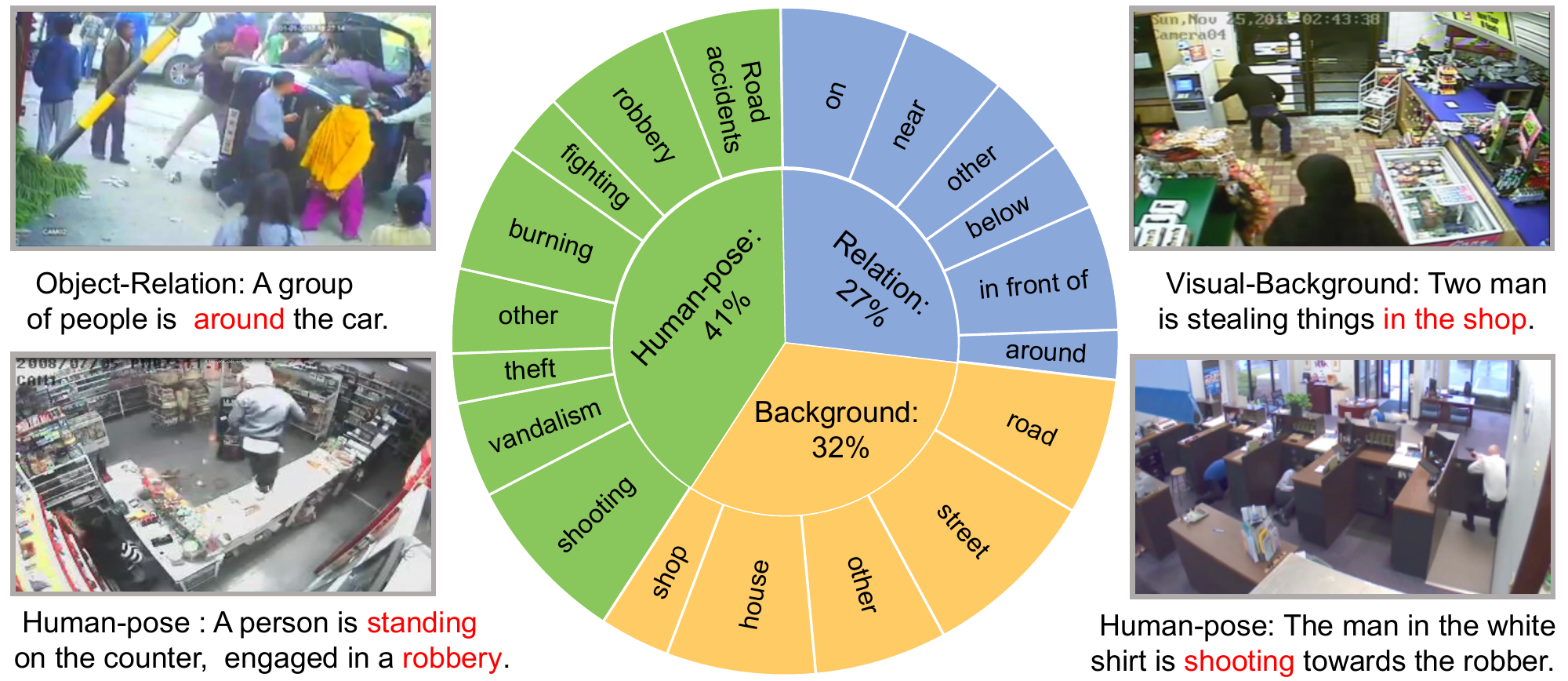} % 调整宽度适应单栏
    \captionsetup{skip=1.5pt} % 图与caption更近（0pt~4pt 之间调）
    \caption{Physical-world information statistics of our dataset. }
    \label{fig:dataset}

\end{figure}

\vspace{-1.5em}

\subsection{Physical-world Trade-off Regularizer}
In this section, we introduce the Physical-world Trade-off Regularizer (PTR) to address the trade-off challenge among the above four experts. PTR consists of a Trade-off Aware Expert Router and a Gated Physical-world Trade-off Loss, detailed as follows:

\textbf{Trade-off Aware Expert Router.} The four experts are regulated by a dynamic \textit{Expert Router} to trade off the coarse-to-fine physical-world information.

Figure \ref{fig:dataset} presents the statistical distribution of physical-world information in our datasets, with coarse-grained and human-pose information constituting the dominant proportion (More detailed dataset distributions are presented in Supplementary Section 1). 
This imbalance in the distribution of physical-world information may cause the model to overemphasize high-proportion categories (e.g., human pose and coarse-grained experts) while neglecting those with lower proportions (e.g., object relations and visual background), as shown in Figure \ref{fig:bar 7} w/o PTR.

The \textit{Expert Router} utilizes an FFN to process input tokens and compute weights for each expert.
% Inspired by \cite{Gou_Liu_Chen_Hong_Xu_Li_Yeung_Kwok_Zhang}, we propose a gated mechanism to trade off information between coarse-grained and fine-grained contexts. The output  $\mathbf{z}$  for all experts is computed as follows:
Different from traditional basic Router \cite{moe3}, we propose a gated mechanism to trade off information between coarse-grained and fine-grained contexts. The output $\mathbf{z}$ for all experts is computed as follows:
\setlength{\abovedisplayskip}{-0.2pt} 
\setlength{\belowdisplayskip}{-0.2pt} 
\begin{equation}
\mathbf{z} = \sum_{i=1}^N \mathbf{R}(h)_i \cdot \mathbf{E}_i(h) + (1 - \mathbf{G}) \mathbf{E}_g(h)
\end{equation}
where $ \mathbf{R}(h)_i $ represents the weight assigned to the \( i \)-th expert, which is used to trade off the fine-grained experts. $ \mathbf{{E}}_i(h) $ denotes the output from the fine-grained experts associated with HPE, ORE, and VBE. \( N \) signifies the total number of fine-grained experts involved. $ \mathbf{E}_g(h) $ corresponds to the output from the coarse-grained expert. Finally, \( \mathbf{G} \) is defined as the maximum element of $ \mathbf{R}(h)_i $, used to better trade off the weighting among coarse-grained and fine-grained experts.

\begin{figure}
    \centering
    \includegraphics[width=0.48\textwidth]{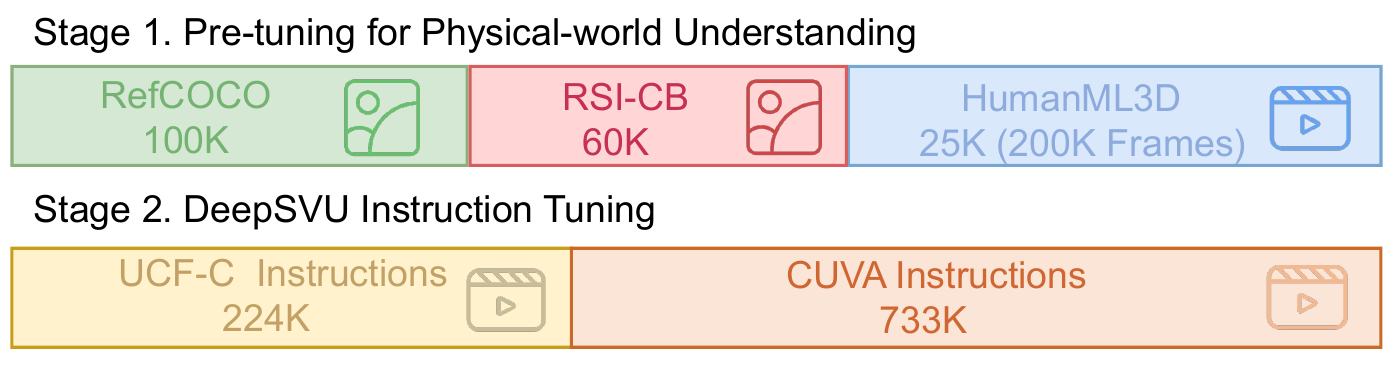}
        \captionsetup{skip=1.5pt} % 图与caption更近（0pt~4pt 之间调）
    % \setlength{\belowcaptionskip}{-2 ex}

    % \caption{\textbf{Training data composition for DeepSVU.} In Stage~1, RefCOCO, HumanML3D, and RSI-CB are used to improve physical-world comprehension. In Stage~2, UCF-C and CUVA instructions are constructed for the DeepSVU task.}
    
    \caption{Data composition for training and inference.}

    \label{fig:4dataset}
\end{figure}

\textbf{Gated Physical-world Trade-off Loss.}   
% To address this issue and maintain a balance among all experts, we design an auxiliary loss function, drawing inspiration from \cite{Zoph_Bello_Kumar_Du_Huang_Dean_Shazeer_Fedus_2022}. The auxiliary loss functions are formulated as follows:
To further trade off among all experts, we design a loss function to penalize imbalanced contributions and regulate the impact of each expert. The loss function is formulated as follows:
\setlength{\abovedisplayskip}{-0.2pt} 
\setlength{\belowdisplayskip}{-0.2pt} 
\begin{equation}
 \mathcal{L}_{\mathit{z}}= \frac{1}{T} \sum_{i=1}^T \left( \log (\sum_{j=1}^N e^{\mathbf{x}^{(i)}_j} +e^{\mathbf{x}_g}) \right)^2
\end{equation}
where $T$ denotes the total number of tokens. 
$N$ represents the number of fine-grained experts.
{$\mathbf{x}_j$} denotes the output of each fine-grained expert, and {$\mathbf{x}_g$} represents the output of the coarse-grained expert. This loss function is specifically designed to penalize large logits. The use of the \(\sum e^{x}\) ensures that scenarios where one or more fine-grained experts produce excessively large outputs are heavily penalized. Additionally, the inclusion of \( e^{x_g} \) in the summation enforces a constraint on the coarse-grained expert, preventing it from dominating the final prediction. By squaring the outputs, the loss function amplifies the penalty for large aggregated logits, making it more sensitive to imbalances in expert contributions.
The total loss for the model can be expressed as: \(\mathcal{L} = \mathcal{L}_{\mathit{ce}}+\alpha \mathcal{L}_{\mathit{z}}\)
% \setlength{\abovedisplayskip}{-0.2pt} 
% \setlength{\belowdisplayskip}{-0.2pt} 
% \begin{equation}
% \mathcal{L} = \mathcal{L}_{\mathit{ce}}+\alpha \mathcal{L}_{\mathit{z}}
% \end{equation}
, where $\mathcal{L}_{ce}$ represents the language modeling loss, $\alpha$ denotes the coefficients of $\mathcal{L}_{z}$ and is tuned to be 0.05.

\subsection{Training strategies}
Video-LLaVA combines ImageBind-based multimodal encoders with the Vicuna language model and performs well on vision instruction-following tasks, but lacks physical-world understanding, which is essential for our study. To address this, we adopt a two-stage training strategy.

\textbf{Stage 1. Pre-tuning for Physical-world Understanding.} As shown in Figure \ref{fig:4dataset}, we pre-tune Video-LLaVA using three high-quality datasets: RefCOCO \cite{lin2014microsoft}, HumanML3D \cite{guo2022generating}, and RSI-CB \cite{RSI-CB}, to strengthen its understanding of the physical-world. For each dataset, we focus on enabling the model to grasp the corresponding physical-world information.

\textbf{Stage 2. DeepSVU Instruction Tuning.} In the second stage, our objective is to enable the model to identify, locate and attribute threatening segments. To support this, we propose the DeepSVU instructions dataset, as detailed in Section \ref{first:fourth_chapter}. 
The pre-trained Video-LLaVA model is instructed to identify, locate, and attribute threats in the video.

%% file: contents/experiments.tex
\begin{table*}[t]
\centering
\setlength{\abovecaptionskip}{-0 ex}
\setlength{\belowcaptionskip}{-0.5 ex}
\caption{Comparison of several Video-LLMs and UPRM  on the Identifying and Locating capability of the CUVA instructions datasets. The $\downarrow$ next to FNRs indicates that a lower metric corresponds to better performance.}
\label{tab:main_results}

\resizebox{\linewidth}{!}{
\tiny
\begin{tabular}{c|cc|cccc|ccccc}
\hline
\toprule[1.2pt]
\multirow{3}{*}{\textbf{Approach}} & \multicolumn{2}{c|}{\textbf{Identifying}}                                  & \multicolumn{4}{c|}{\textbf{Locating}}                                                            & \multicolumn{5}{c}{\textbf{Attributing}}                                                                                                                                                              \\ \cline{2-12}
                          & \multicolumn{1}{c|}{\multirow{2}{*}{FNRs$\downarrow$}} & \multirow{2}{*}{F2} &  \multicolumn{4}{c|}{mAP@tIoU}                                                                   & \multicolumn{1}{c}{\multirow{2}{*}{SB}} & \multicolumn{1}{c}{\multirow{2}{*}{Rouge}} & \multicolumn{1}{c}{\multirow{2}{*}{BLEU}} & \multicolumn{1}{c}{\multirow{2}{*}{GPT}} & \multirow{2}{*}{Human}         \\ \cline{5-6}
                          & \multicolumn{1}{c|}{}                     &                     & \multicolumn{1}{c}{0.1} & \multicolumn{1}{c}{0.3} & \multicolumn{1}{c}{0.5} & \multicolumn{1}{c|}{Average} & \multicolumn{1}{c}{}                   & \multicolumn{1}{c}{}                   & \multicolumn{1}{c}{}                   & \multicolumn{1}{c}{}                   &                         \\ \hline
Video Chat             & \multicolumn{1}{c|}{\textcolor{blue}{32.65}}  & \multicolumn{1}{c|}{69.93}              & 90.55 & 78.61 & 58.71 & \multicolumn{1}{c|}{75.95}         & 0.39 & 0.36  & 27.94 & 6.54 & 3.23                   \\ 
% Video ChatGPT          & \multicolumn{1}{c|}{\textcolor{blue}{39.09}}   & \multicolumn{1}{c|}{62.65}             & 93.03  & 78.61 & 51.24 & \multicolumn{1}{c|}{74.29}          & 0.44  & 0.4  & 32.44 & 6.85 &3.52                 \\
Valley                 & \multicolumn{1}{c|}{\textcolor{blue}{58.91}}    & \multicolumn{1}{c|}{45}            & 92.04 & 67.66 & 14.43 & \multicolumn{1}{c|}{58.04}        & 0.3  & 0.31 & 22.1 & 5.93 & 2.97                \\
PandaGPT              & \multicolumn{1}{c|}{\textcolor{blue}{23.36}}   & \multicolumn{1}{c|}{78.32}             & 93.53 & 81.59 & 70.15 & 81.76                & 0.44  & 0.40  & 35.51 &7.06 & 3.61             \\
mPLUG-Owl              & \multicolumn{1}{c|}{\textcolor{blue}{42.35}}    & \multicolumn{1}{c|}{60.19}            & 89.55 & 67.66 & 48.76 & \multicolumn{1}{c|}{68.66}     & 0.42 & 0.33  & 29.46 & 6.04 & 3.06                  \\
Chat-UniVi            & \multicolumn{1}{c|}{\textcolor{blue}{38.66}}     & \multicolumn{1}{c|}{61.46}           & 88.51 & 78.11 & 39.3 & 68.64               & 0.41  & 0.34  & 31.33 & 6.23 & 3.18               \\
Video-LLaVA            & \multicolumn{1}{c|}{\textcolor{blue}{25.17}}    & \multicolumn{1}{c|}{74.37}            & 90.05 & 72.64 & 58.21 & \multicolumn{1}{c|}{73.63}          & 0.4  & 0.37  & 28.51 & 6.69 & 3.49                 \\ 
Omni-SILA    & \multicolumn{1}{c|}{\textcolor{blue}{14.32}}    & \multicolumn{1}{c|}{83.55}            & 91.7 & 80.35 & 69.32 & \multicolumn{1}{c|}{80.4}          & 0.47  & 0.42  & 35.13 & 7.35 & 4.01                 \\
Holmes            & \multicolumn{1}{c|}{\textcolor{blue}{23.78}}    & \multicolumn{1}{c|}{79.77}            & 90.57 & 81.63 & 67.34 & \multicolumn{1}{c|}{79.85}          & 0.49  & 0.38  & 25.43 & 6.75 & 3.63                 \\
Hawkeye    & \multicolumn{1}{c|}{\textcolor{blue}{13.24}}    & \multicolumn{1}{c|}{84.35}            & 92.17 & 83.37 & 70.03 & \multicolumn{1}{c|}{81.85}          & 0.47  & 0.43  & 35.64 & 7.32 & 4.06                 \\

\hline

\rowcolor{lightpink}UPRM                & \multicolumn{1}{c|}{\textbf{\textcolor{blue}{10.46}}}    & \multicolumn{1}{c|}{\textbf{88.41}}            & \textbf{93.03} & \textbf{86.07} & \textbf{78.11} & \multicolumn{1}{c|}{\textbf{86.24}}            & \textbf{0.51}  &  \textbf{0.46}  & \textbf{42.15} & \textbf{7.51} & \textbf{4.21}        \\

\rowcolor{lightblue} w/o HPE       & \multicolumn{1}{c|}{\textcolor{blue}{15.16}}    & \multicolumn{1}{c|}{79.13}            & 88.13 & 82.54 & 66.67 & \multicolumn{1}{c|}{79.11}           & 0.46 & 0.4  & 30.18 & 7.01 & 4.01                  \\

\rowcolor{lightblue} w/o ORE      & \multicolumn{1}{c|}{\textcolor{blue}{11.41}}      & \multicolumn{1}{c|}{86.64}          & 93.53 & 84.08 & 74.13 & \multicolumn{1}{c|}{83.91}             & 0.46  & 0.44  & 40.51 & 7.37 & 3.98               \\
 
\rowcolor{lightblue} w/o VBE      & \multicolumn{1}{c|}{\textcolor{blue}{11.17}}      & \multicolumn{1}{c|}{88.15}          & 94.53 & 85.07 & 76.12 & \multicolumn{1}{c|}{85.24}                & 0.49 & 0.45  & 41.81 & 7.44 & 4.13            \\

\rowcolor{lightorange} w/o UPE       & \multicolumn{1}{c|}{\textcolor{blue}{19.49}}    & \multicolumn{1}{c|}{81.13}            & 89.13  & 73.03 & 63.38 & 75.18         & 0.41   & 0.38  & 32.14 & 6.94 & 3.65                 \\
\rowcolor{lightorange} w/o PTR        & \multicolumn{1}{c|}{\textcolor{blue}{18.07}}    & \multicolumn{1}{c|}{83.25}            & 88.5 & 82.04 & 71.08 & \multicolumn{1}{c|}{80.54}              & 0.47 & 0.41  & 37.56 & 7.15 & 3.77              \\
\rowcolor{lightorange} w/o Pre-tuning        & \multicolumn{1}{c|}{\textcolor{blue}{21.16}}    & \multicolumn{1}{c|}{80.43}            & 89.05 & 76.62 & 62.19 & \multicolumn{1}{c|}{75.95}           & 0.45   & 0.4  & 30.21 & 6.74 & 3.61               \\
 
\bottomrule[1.2pt]   

\end{tabular}
}
\end{table*}

\begin{table}[]

\resizebox{\linewidth}{!}{
% \tiny
\begin{tabular}{c|cc|cccc}
% \hline
\specialrule{1.2pt}{0pt}{0pt}
& \multicolumn{2}{c|}{\textbf{Identifying}} & \multicolumn{4}{c}{\textbf{Locating}}                                                  \\ \cline{2-7} 
                           &       
                           &              &               & \multicolumn{2}{c}{mAP@tIoU} &            \\ \cline{5-6}
\multirow{-3}{*}{\textbf{Models}} & \multirow{-2}{*}{FNRs$\downarrow$}                & \multirow{-2}{*}{F2}        & 0.1                                   & 0.2                                   & 0.3                                   & \multirow{-2}{*}{Average}              \\ \hline

VadClip\textsuperscript{†}            & \multicolumn{1}{c|}{\textcolor{blue}{67.54}}    & \multicolumn{1}{c|}{31.62}              & 22.69 & 13.03 & 7.98 & 14.57                                \\
BiConvLSTM\textsuperscript{†}             & \multicolumn{1}{c|}{\textcolor{blue}{81.99}}    & \multicolumn{1}{c|}{18.95}              & 21.32 & 8.46 & 4.78 & 11.52      \\ 
X3D\textsuperscript{†}             & \multicolumn{1}{c|}{\textcolor{blue}{77.2}}    & \multicolumn{1}{c|}{23.16}              & 18.68 & 10.89 & 5.84 & 11.8                     \\
CLIP-TSA\textsuperscript{†}             & \multicolumn{1}{c|}{\textcolor{blue}{65.05}}    & \multicolumn{1}{c|}{31.5}              & 22.22 & 12.26 & 4.98 & 13.15                 \\

Video Chat             & \multicolumn{1}{c|}{\textcolor{blue}{61.4}}    & \multicolumn{1}{c|}{32.27}              & 29.11 & 14.77 & 6.75 & 16.88                          \\

% Video ChatGPT          & \multicolumn{1}{c|}{\textcolor{blue}{78.48}}   & \multicolumn{1}{c|}{17.99}             & 17.02 & 6.83 & 4.25 & 9.37                          \\
Valley                  & \multicolumn{1}{c|}{\textcolor{blue}{68.6}}    & \multicolumn{1}{c|}{27.51}            & 28.79 & 14.14 & 7.73 & 16.87                        \\
PandaGPT              & \multicolumn{1}{c|}{\textcolor{blue}{68.77}}   & \multicolumn{1}{c|}{21.72}             & 18.65 & 9.52 & 4.76 & 10.98                    \\
mPLUG-Owl              & \multicolumn{1}{c|}{\textcolor{blue}{54.13}}   & \multicolumn{1}{c|}{32.36}             & 29.66 & 16.55 & 6.9 & 17.7                        \\
Chat-UniVi            & \multicolumn{1}{c|}{\textcolor{blue}{80.52}}     & \multicolumn{1}{c|}{21.65}           & 16.9 & 7.75 & 2.11 & 8.92                       \\
Video-LLaVA            & \multicolumn{1}{c|}{\textcolor{blue}{76.34}}   & \multicolumn{1}{c|}{23.65}             & 19.01 & 9.38 & 4.57 & 10.99                        \\
Omni-SILA & \multicolumn{1}{c|}{\textcolor{blue}{44.36}}   & \multicolumn{1}{c|}{42.95}              & 32.36 & 19.34 & 11.65 & 21.11   
        \\
Holmes            & \multicolumn{1}{c|}{\textcolor{blue}{42.37}}   & \multicolumn{1}{c|}{45.5}              & 34.97 & 21.33 & 13.99 & 23.43                      \\
Hawkeye & \multicolumn{1}{c|}{\textcolor{blue}{46.74}}   & \multicolumn{1}{c|}{44.13}              & 33.17 & 17.84 & 13.2 & 21.4                      \\

\hline

\rowcolor{lightpink}UPRM                & \multicolumn{1}{c|}{\textbf{\textcolor{blue}{37.28}}}    & \multicolumn{1}{c|}{\textbf{49.33}}            & \textbf{39.72} & \textbf{23.34} & \textbf{15.68} & \textbf{26.25}                 \\

\rowcolor{lightblue} w/o HPE       & \multicolumn{1}{c|}{\textcolor{blue}{43.19}}    & \multicolumn{1}{c|}{43.04}            & 33.57 & 21.33 & 13.64 & 22.84                                  \\

\rowcolor{lightblue} w/o ORE      & \multicolumn{1}{c|}{\textcolor{blue}{38.17}}      & \multicolumn{1}{c|}{46.09}          & 31.36 & 21.25 & 13.59 & 22.07                        \\

\rowcolor{lightblue} w/o VBE      & \multicolumn{1}{c|}{\textcolor{blue}{41.75}}      & \multicolumn{1}{c|}{42.56}          & 29.24 & 17.33 & 9.39 & 18.65                          \\

\rowcolor{lightorange} w/o UPE       & \multicolumn{1}{c|}{\textcolor{blue}{53.37}}    & \multicolumn{1}{c|}{40.58}            & 29.37 & 17.06 & 9.13 & 18.52                                    \\
\rowcolor{lightorange} w/o PTR        & \multicolumn{1}{c|}{\textcolor{blue}{44.84}}   & \multicolumn{1}{c|}{44.07}             & 38.19 & 19.1 & 9.72 & 22.34                                \\
\rowcolor{lightorange} w/o Pre-tuning        & \multicolumn{1}{c|}{\textcolor{blue}{54.2}}   & \multicolumn{1}{c|}{36.31}             & 29.66 & 16.55 & 6.9 & 17.7           \\
\specialrule{1.2pt}{0pt}{0pt}
\end{tabular}}
\caption{Comparison of Identifying and Locating capabilities among several well-performing Video-LLMs and traditional non-LLM approaches conducted on the UCF-C instructions datasets, where \textsuperscript{†} denotes non-LLM approaches.}
\label{tab:2}
\end{table} 

\section{EXPERIMENTAL SETTINGS}

% ######4.1
\subsection{Instructions Dataset Construction}\label{first:fourth_chapter}

The existing \textbf{Video-LLaVA} framework shows limited physical-world comprehension capabilities. To overcome this, we propose a two-stage training method designed to enhance the model's physical-world reasoning abilities. The following sections outline the construction of instructions datasets for each stage.

%55555
% 传统异常检测比较：
\begin{table}[] % 强制定位当前表格
% % 保存原始参数值
% \newlength{\oldtextfloatsep}
% \setlength{\oldtextfloatsep}{\textfloatsep}
% \newlength{\oldintextsep}
% \setlength{\oldintextsep}{\intextsep}
% \newlength{\oldskip}
% \setlength{\oldskip}{\belowcaptionskip}

% % 局部参数调整
% \setlength{\textfloatsep}{0pt}   % 仅当前表格与正文间距
% \setlength{\intextsep}{0pt}      % 仅当前表格顶部/底部间距
% \captionsetup[table]{skip=0pt}   % 仅当前标题间距

\setlength{\belowcaptionskip}{2 ex}
\resizebox{\linewidth}{!}{
% \tiny
\begin{tabular}{c|cc|cccc}
% \hline
\specialrule{1.2pt}{0pt}{0pt}
& \multicolumn{2}{c|}{\textbf{Identifying}} & \multicolumn{4}{c}{\textbf{Locating}}                                                   \\ \cline{2-7} 
                           &       
                           &              &               & \multicolumn{2}{c}{mAP@tIoU} &            \\ \cline{5-6}
\multirow{-3}{*}{\textbf{Models}} & \multirow{-2}{*}{FNRs$\downarrow$}                & \multirow{-2}{*}{F2}        & 0.1                                   & 0.3                                   & 0.5                                   & \multirow{-2}{*}{Average}             \\ \hline

VadClip             & \multicolumn{1}{c|}{\textcolor{blue}{63.29}}    & \multicolumn{1}{c|}{37.44}              & 87.06 & 39.4 & 13.12 & 46.53                                \\
BiConvLSTM            & \multicolumn{1}{c|}{\textcolor{blue}{62.1}}    & \multicolumn{1}{c|}{42.02}              & 81.59 & 43.78 & 19.4 & 48.26      \\ 
X3D            & \multicolumn{1}{c|}{\textcolor{blue}{43.96}}    & \multicolumn{1}{c|}{58.74}              & 89.55 & 64.18 & 45.77 & 66.5                     \\
CLIP-TSA            & \multicolumn{1}{c|}{\textcolor{blue}{61.45}}    & \multicolumn{1}{c|}{46.22}              & 84.12 & 49.56 & 15.72 & 49.8                  \\

\hline

\rowcolor{lightpink}UPRM                & \multicolumn{1}{c|}{\textbf{\textcolor{blue}{10.46}}}    & \multicolumn{1}{c|}{\textbf{88.41}}            & \textbf{93.03} & \textbf{86.07} & \textbf{78.11} & \textbf{86.24}                                  \\

\specialrule{1.2pt}{0pt}{0pt}
\end{tabular}}
\caption{Comparison of Identifying and Locating capabilities among several well-performing non-LLM threat detection models on the CUVA instructions datasets.}
\label{tab:cuva-tradition}

% % 恢复全局参数（关键！）
% \setlength{\textfloatsep}{\oldtextfloatsep}
% \setlength{\intextsep}{\oldintextsep}
% \setlength{\belowcaptionskip}{\oldskip}
\end{table}

\textbf{Stage 1.} To enhance the physical-world comprehension ability of \textbf{UPRM}, we construct a new dataset by integrating samples from RefCOCO, HumanML3D, and RSI-CB. Physical-world specific instructions are manually designed for each dataset to align with the training objectives. For instance: ``\textit{Instruction: Analyze the designated area \textless region\textgreater and the actions depicted in the video. Response: A woman is dressed in a red dress and approaches a car to open its door.}" Here, \textit{\textless region\textgreater}  specifies the target spatial coordinates. To ensure dataset balance, we carefully select subsets from the above datasets, resulting in 25,000 videos and 160,000 images.

\textbf{Stage 2.} We construct two DeepSVU instruction datasets, i.e., UCF-C instructions and CUVA instructions, based on two widely-used anomaly detection datasets UCF-Crime \cite{ucf-crime} and CUVA \cite{cuva}, and treat the anomaly categories (e.g., shootings, robberies) as the threat types in our task. Specifically, the CUVA instructions dataset is built based on the public CUVA \cite{cuva}, which provides the root cause of the threat in the video. Then, we use this dataset to evaluate the model's capabilities in \textbf{identifying}, \textbf{locating}, and \textbf{attributing} threat. However, the original dataset's instructions cannot support our task, as they only include video captions of threats, so we reconstruct the instructions to better align with our three threat identifying, locating and attributing sub-tasks (as shown in Figure 1). Specifically, we construct the corresponding instructions for asking the model to identify ``\textit{Please determine whether there exist threats  in the video}”, locate ``\textit{Please locate the exact timestamps where the threats occur}”, and attribute ``\textit{Please give me a detailed explanation of the root causes of the threat}”. The corresponding responses are: ``\textit{Yes, there exists a threat}”, ``\textit{The threat occurs between \textless t1\textgreater  and \textless t2\textgreater,}” and ``\textit{The threat is attributed to...,}”. Further,
our UCF-C instructions dataset is built based on the public UCF-Crime \cite{ucf-crime}. Since the original dataset does not provide explanations for the threats, we use it to evaluate the model's capabilities in \textbf{identifying} and \textbf{locating} threats, and reconstruct the same identifying and locating instructions like our CUVA instructions dataset. 

 \begin{figure}[t!]
    \centering
    \hspace{-0.5cm}
    \includegraphics[width=0.5\textwidth, scale=0.39]{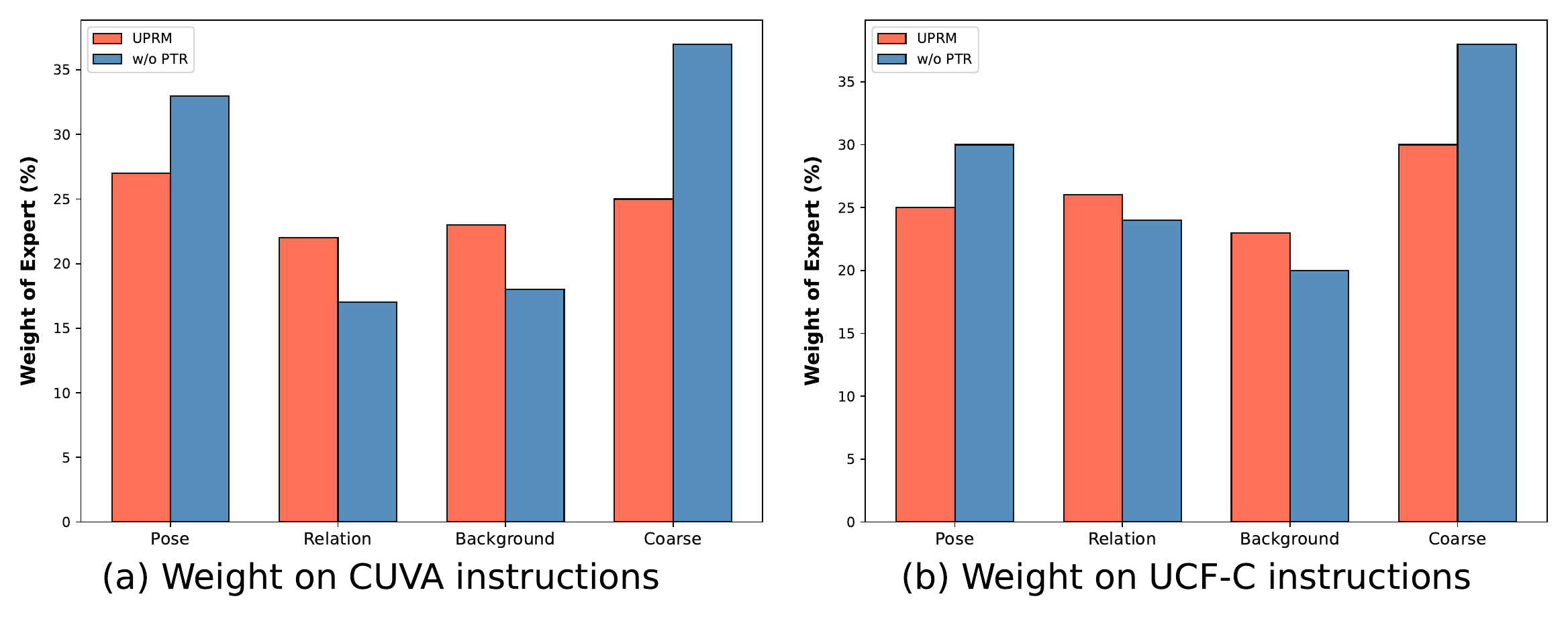}
    \setlength{\abovecaptionskip}{-3 ex}
    \caption{Comparisons of Weights Assigned by Different Experts. w/o PTR is exactly the basic MoE expert weighting.}
    \label{fig:bar 7}
\end{figure}

\subsection{Baselines and Implementation Details}
We select several advanced Video-LLMs to serve as baselines for the analysis: \textbf{VideoChat} \cite{videochat}, \textbf{Valley} \cite{valley}, \textbf{mPLUG-Owl} \cite{ye2023mplug}, PandaGPT \cite{pandagpt}, \textbf{Chat-UniVi} \cite{chatunivi}, \textbf{OmniSILA}\cite{luo2025omnisila}, \textbf{Holmes} \cite{zhang2024holmes}, and \textbf{Hawkeye} \cite{jinyujiehawkeye}. All Video-LLMs use 7B parameter configurations. For the non-LLM baseline approaches, we selected \textbf{VadClip} \cite{vadclip}, \textbf{BiConvLSTM} \cite{BiConvLSTM}, \textbf{X3D} \cite{x3d} and \textbf{CLIP-TSA} \cite{clip-tsa} to evaluate the abilities of identifying and locating.

Since the above baseline models focus on different tasks and employ different experimental settings, for a fair and thorough comparison, we reimplement these models and leverage their released codes to obtain experimental results on our UCF-C instruction and CUVA instruction datasets. The hyperparameters of these baselines reported by their public papers still adopt the same setting. The others and the hyperparameters of our UPRM approach is tuned according to the best performance. Specifically, for  UPRM approach, training is initiated using the AdamW optimizer  with a learning rate of 2e-5.       A warm-up strategy is applied, with a warm-up ratio set to 0.03. Subsequently, we fine-tune Video-LLaVA (7B) using LoRA \cite{lora}. The LoRA matrix is configured with a dimension of 16, a scaling factor of 32, and a dropout rate of 0.05. The batch size is set to 8, with training for one epoch.  During all training phases, the parameters of HigherHRNet, RelTR, and SAM are kept frozen to prevent overfitting and preserve their pre-trained functionality.  
All approaches are reproduced with PyTorch in a consistent setup, including NVIDIA A100-PCIE-40GB, Intel Xeon Gold 6248R CPU @ 3.00GHz, CUDA version 12.2, PyTorch 2.0.1, and Python 3.10.9.

\subsection{Evaluation Metrics}
For the DeepSVU task, which aims to precisely identify, locate, and attribute threats, we adopt several key evaluation metrics. 

\begin{figure} 
    \centering
    \setlength{\abovecaptionskip}{-0.1 ex}
    \includegraphics[width=0.235\textwidth]{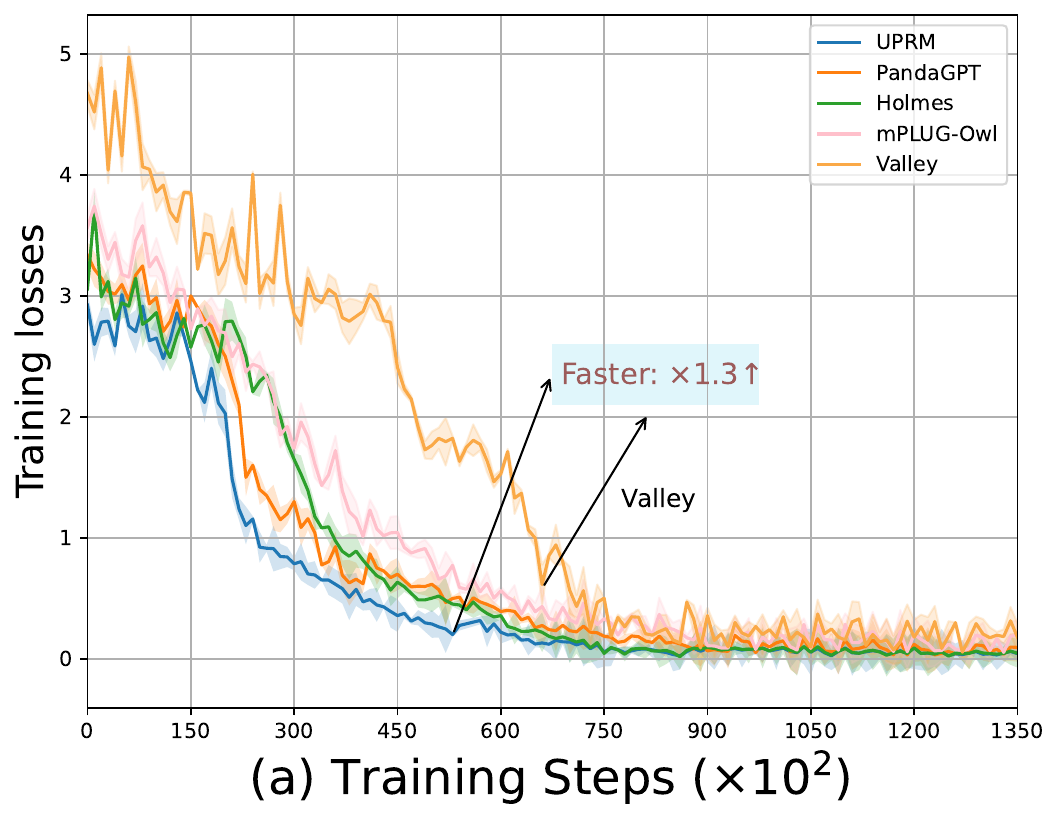}
      \hfill%
    \includegraphics[width=0.235\textwidth]{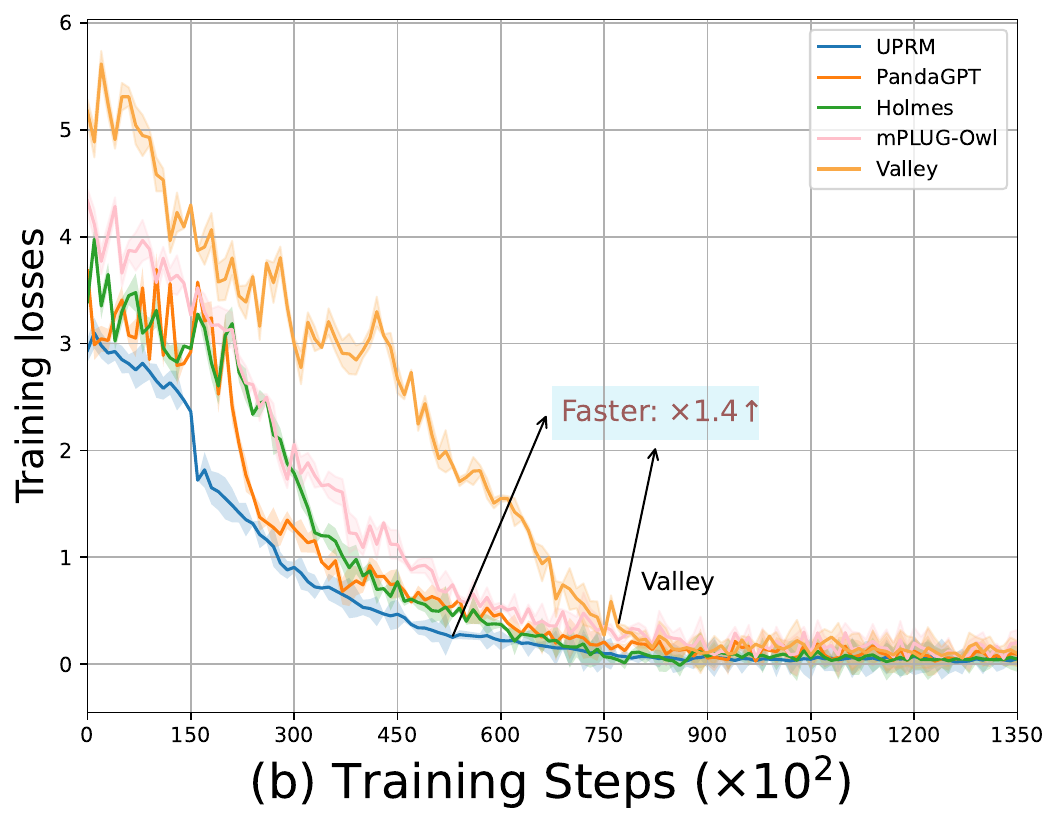}
    
    \caption{Convergence analysis of UPRM and other Video-LLMs on (a) CUVA and (b) UCF-C instruction datasets.}

        % \caption{\textbf{Convergence Analysis.} Comparison of the UPRM model with other Video-LLMs through training loss trajectories on (a) CUVA and (b) UCF-C instructions datasets.}
 
   \label{fig:loss}
\end{figure}

% This approach is critical for evaluating the precision of locating performance \cite{TSL}.

For threat identifying, DeepSVU prioritizes minimizing missed detections of threatening frames. Following recent works \cite{jinyujiehawkeye}, we emphasize the importance of reducing False-Negative Rates (FNRs), which measure the likelihood of erroneously classifying a threat frame as normal. This metric is crucial, especially considering scenarios where overlooking a threat could lead to severe consequences. Besides FNRs, the F2-score with a bias towards recall is used, reflecting our preference for Recall over Precision. 

For threat locating, we use the metric of mean Average Precision (mAP) across varying Intersection over Union (IoU) thresholds, denoted as mAP@IoU. For the UCF-C Instructions dataset, the IoU thresholds are set to 0.1, 0.2, and 0.3. For the CUVA Instructions dataset, the IoU thresholds are set to 0.1, 0.3, and 0.5.
% The equations representing these metrics are given by:
% {\begin{align}
%     \text{FNRs} = \frac{\alpha \text{ of false-negative frame}}{\alpha \text{ of positive frame}}\\
%     F_{\beta} = \frac{(1 + \beta^2) \times \text{Precision} \times \text{Recall}}{\beta^2 \times \text{Precision} + \text{Recall}}
% \end{align}
% }
% where \text{$\alpha$} denotes the frame numbers and $\beta$ is 2. 

For threat attributing, prior approaches lack the capability to effectively attribute and detailly evaluate the threat causes. Thus, we take advantage of three widely-used generation metrics to detailly assess the causes, i.e., ROUGE, Sentence-BERT (SB), and BLEU.  First, ROUGE measures lexical overlap between the generated attribution and the true cause, offering a surface-level alignment score. 
Sentence-BERT captures semantic similarity between the generated attribution and the ground truth. BLEU further evaluates surface-level alignment with an emphasis on fluency and precision.
The rationality of attributions is assessed using both automated and human evaluation approaches.
For automated evaluation, GPT-based scoring on a 1–10 scale measures two criteria: cause relevance and consistency with sentiment clues.
Human evaluation involves three annotators who rate rationality on a 1–5 scale using predefined guidelines.

%% file: contents/analysis.tex
% 专属补偿（仅作用于当前表格）

% \vspace*{-2\baselineskip} % 关键补偿（根据效果微调数值）

\begin{figure*}
    \centering
    \vspace{3mm}
    \includegraphics[width=\textwidth, scale=0.39]{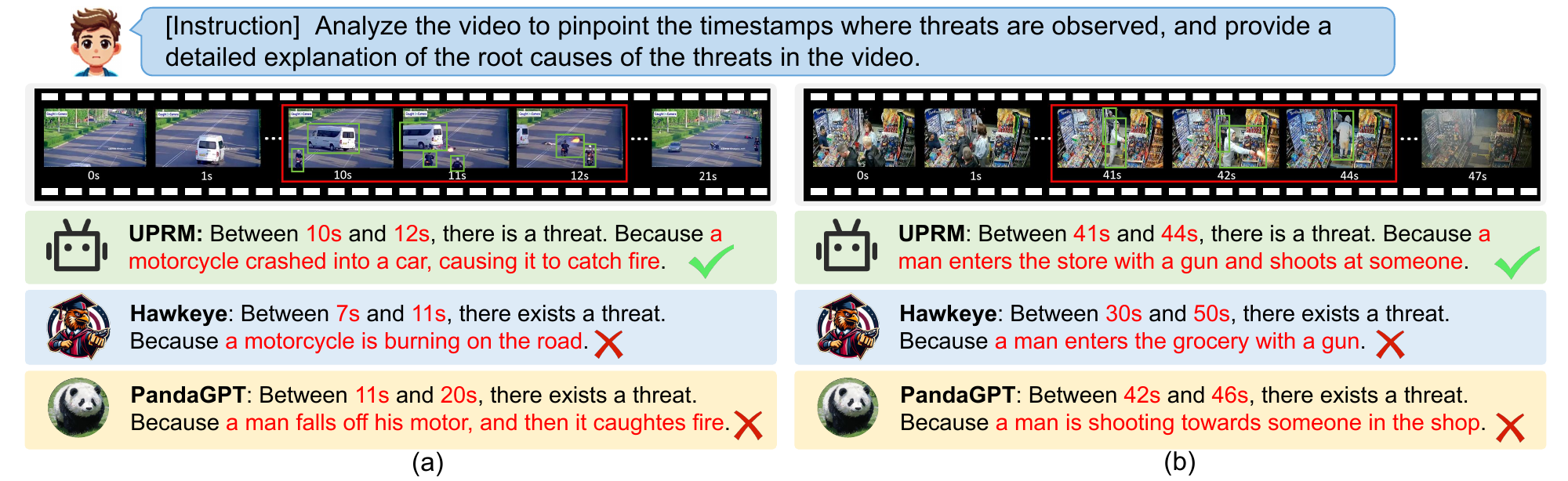}
    \setlength{\abovecaptionskip}{-3 ex}
    \setlength{\belowcaptionskip}{-2 ex}
    \caption{\textbf{Two samples to compare UPRM with other Video-LLMs.} }
    \label{fig:casestudy}
\end{figure*}

\section{Results and Discussions}
% We evaluate the performance of the UPRM  in comparison to several non-LLM VAD approachs and Video-LLMs. 

\subsection{Main Results}
\textbf{Threat Identifying and Locating.} As demonstrated in Tables \ref{tab:main_results}, \ref{tab:2}, and \ref{tab:cuva-tradition}, UPRM outperforms both traditional non-LLM SVU approaches and other Video-LLMs baselines. Specifically, on the CUVA instructions dataset, it outperforms the best non-LLM baseline (i.e., X3D) with a 33.5 reduction in FNRs and a 29.67 improvement in F2 score, in addition to a 19.74 increase in average mAP@tIoU. Furthermore, UPRM surpasses best Video-LLM (i.e., Hawkeye) with a 2.78 reduction in FNRs and a 4.06 enhancement in F2 score, alongside a 4.39 improvement in average mAP@tIoU. On the UCF-C instructions dataset, UPRM exceeds the best non-LLM (i.e., CLIP-TSA) baseline by reducing FNRs by 27.77, improving the F2 score by 17.83, and achieving a 11.68 boost in average mAP@tIoU. It also outperforms the best Video-LLM (i.e., Holmes) with a 5.09 reduction in FNRs and a 3.83 improvement in F2, along with a 2.82 increase in mAP@tIoU.

\textbf{Threat Attributing.} Table \ref{tab:main_results} shows that UPRM outperforms all Video-LLMs, surpassing the best Video-LLM (i.e., Hawkeye) by improvements of 0.04 in SB, 0.03 in ROUGE, 6.51 in BLEU, 0.19 in GPT, and 0.15 in Human score.

% \begin{figure*}\setcounter{figure}{0}
%     \includegraphics[width=0.5\textwidth]{figure/heat1.pdf}
%     \includegraphics[width=0.5\textwidth]{figure/heat2.pdf}
     
%  \caption{Weight of Expert.}
%   \label{fig:heat}
%   \end{figure*}

% \begin{figure} 
%     \centering
%     \includegraphics[width=0.48\textwidth]{figure/heat2.png} % Removed scale, kept width
%     \setlength{\abovecaptionskip}{-2ex} % Ensuring ex units are used properly
%     \setlength{\belowcaptionskip}{-3ex}
%     \caption{Weight of Expert.}
%     \label{fig:heat}
% \end{figure}

\subsection{Ablation Study}
UPRM's performance highlights the importance of specific components, with their impact analyzed in Tables \ref{tab:main_results} and \ref{tab:2}.

\textbf{Effectiveness Study of  Physical-World Information.}
The performance of w/o HPE, w/o ORE, and w/o VBE degrades across all metrics. The performance on CUVA instructions dataset shows a significant decline across all metrics, with an average decrease of 2.12, 3.77, 3.48 in FNRs, F2, mAP@tIoU. Additionally, metrics such as SB, ROUGE, BLEU, GPT, and Human scores have decreased by 0.04, 0.03, 4.65, 0.23, and 0.51, respectively. On UCF-C instructions dataset, similar trends are observed with reductions. \textbf{These results highlight the  effectiveness of components in UPRM.}

\textbf{Effectiveness Study of MoE Block.}
\textbf{1) w/o UPE:} Performance significantly degrades on both datasets. On UCF-C, FNRs increase by 16.09, F2 drops by 8.75, and mAP@tIoU decreases across thresholds. For CUVA instructions, FNRs increase by 9.03, F2 drops by 7.28, mAP@tIoU drops by 11.06, and ROUGE, SB, BLEU, GPT, and human scores drop by 0.08, 0.05, 1.64, 0.57, and 0.56, respectively. \textbf{Removing UPE compromises the model’s robustness and accuracy, underscoring its importance in adapting to varied threats while maintaining high metrics}. \textbf{2) w/o PTR:} It leads to a notable decline in performance. On UCF-C, FNRs rise by 7.56, F2 drops to 5.26, and mAP@tIoU decreases across thresholds. For CUVA instructions, FNRs are 18.07, and evaluation scores also drop significantly. Furthermore, as shown in Figure \ref{fig:bar 7}, before using PTR, the model tends to assign more weight to coarse-grained experts, as introduced in the instructions. However, it can be seen that our method effectively balances the contributions among both experts.

% \begin{figure*} 
%     \centering
%     \includegraphics[width=0.24\textwidth]{figure/a.pdf}
%     \includegraphics[width=0.24\textwidth]{figure/b.pdf}
%     \includegraphics[width=0.24\textwidth]{figure/c.pdf}
%     \includegraphics[width=0.24\textwidth]{figure/d.pdf}
    
%     \caption{\textbf{Convergence Analysis.} (a) Comparison of the UPRM model with other Video-LLMs through training loss trajectories on CUVA instructions. (b) Analysis of the effect of component removal within the UPRM model on CUVA instructions, illustrating the differences in training losses when specific components are excluded. (c) Training loss trajectories of the UPRM model compared to other Video-LLMs on the UCF-C instructions dataset. (d) Examination of the impact of component removal within the UPRM model on UCF-C instructions, highlighting the variations in training losses when certain components are omitted. ``Faster" refers to how many times faster our method's convergence point is compared to the slowest baseline.}
%    \label{fig:loss}
% \end{figure*}

\textbf{Effectiveness Study of Pre-tuning.} The model \textbf{w/o Pre-tuning} shows a significant decline compared to UPRM. On the CUVA instructions dataset, FNRs, F2, and mAP@tIoU decrease by 10.7, 7.98, and 10.29, respectively. Additionally, ROUGE, SB, BLEU, GPT, and human evaluation scores drop to 0.06, 0.06, 11.94, 0.77, and 0.6. The same trend is observed on the UCF-C instructions dataset. 

\subsection{Convergence Analysis of UPRM}
In Figure \ref{fig:loss}, we compare the convergence speed of UPRM with several strong Video-LLMs and its variants without specific components at different training steps. UPRM shows the fastest convergence among all Video-LLMs. At the convergence point, UPRM's loss values are 0.19 on CUVA instructions and 0.23 on UCF-C instructions, highlighting its superior training efficiency. (Supplementary Section 2 includes further convergence analysis and inference time comparisons; Section 3 details fine-grained FNRs robustness.)

\subsection{Qualitative Analysis}

Figure \ref{fig:casestudy} shows a comparison of UPRM with other Video-LLMs. Two samples are randomly selected from both the UCF-C instructions and CUVA instructions datasets, and the models are tasked with identifying threat timestamps and explaining root causes.
In the first case from Figure  \ref{fig:casestudy}(a) (motorcycle accident), UPRM precisely identifies the threat between 10 and 12 seconds, detecting both the collision and fire. In contrast, Hawkeye fails to locate the exact timestamp, and PandaGPT not only misidentifies the timing but also fails to attribute precise causes.
In the second case Figure \ref{fig:casestudy}(b) (gunman entering store), UPRM precisely detects the threat and provides a correct explanation. Similar to the first case, both Hawkeye and PandaGPT locate a broader range and offer less specific explanations. \textbf{These examples demonstrate UPRM’s superior precision in identifying, locating, and attributing threats.}

% \begin{figure} % 这里改为单栏浮动体
%     \centering
%     \includegraphics[width=1\linewidth]{figure/casestudy.pdf}
%     \setlength{\abovecaptionskip}{-1.5 ex}
%     \setlength{\belowcaptionskip}{-0.5 ex}
%     \caption{\textbf{Two samples to compare UPRM with other Video-LLMs.} Our UPRM method outperforms the others, demonstrating superior accuracy in handling video anomaly identifying, locating and attributing tasks.}
%     \label{fig:casestudy}
% \end{figure}

%% file: contents/conclusion.tex
\section{Conclusion}
% In this paper, we introduce a new task, \textbf{I}nteractable Video \textbf{A}nomaly  \textbf{i}dentifying, \textbf{l}ocating and \textbf{a}ttributing  (Inter-Aila), aiming to revolutionize the field of video anomaly detection through the application of Video-LLMs. We presented the cutting-edge UPRM approach, which enhances anomaly detection capabilities in surveillance videos by incorporating detailed Physical-world Enhanced components. Central to our approach are the Fine-grained Human-Pose Expert, Fine-grained Object-Relation Expert, Fine-grained Visual-Background Expert, and Coarse-grained Video Expert, which collectively refine the model's sensitivity to contextual discrepancies in video scenes.
In this paper, we propose an UPRM model for the DeepSVU task, which designs a Physical-world Enhanced
MoE Block and a Physical-world Trade-off Regularizer to address physical-world modeling and trade-off challenges. Extensive experiments, performed on the newly constructed datasets, demonstrate UPRM's superior ability to identify, locate, and attribute threats when compared to several advanced Video-LLMs and non-LLM approaches. In our future work, we would like to explore the integration of additional real-world information, such as events, locations \cite{ligeoreasoner}, and semantic mappings \cite{raychaudhuri2025semantic}, to further enhance the model's predictive accuracy. 
% Moreover,  inference time remains a significant challenge, prompting us to consider lightweight model optimizations, such as LLM-oriented distillation and quantization pruning, to enhance operational efficiency without compromising detection quality.

%% file: contents/supplementary.tex
% =========================================================
% Appendix
% =========================================================

\appendix

\begin{figure*}[t]
  \centering
  \includegraphics[width=0.45\textwidth]{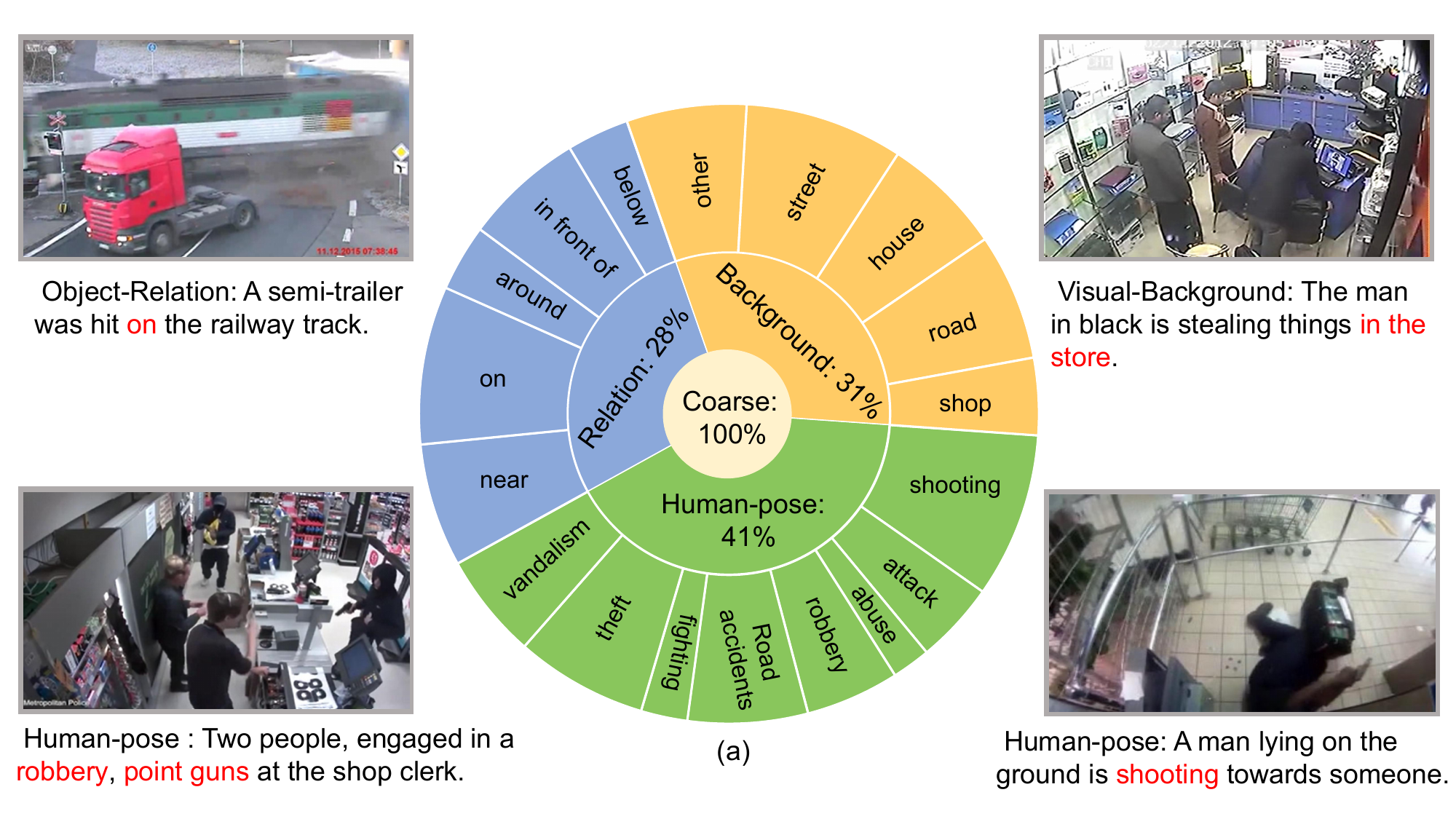}
  \hfill
  \includegraphics[width=0.45\textwidth]{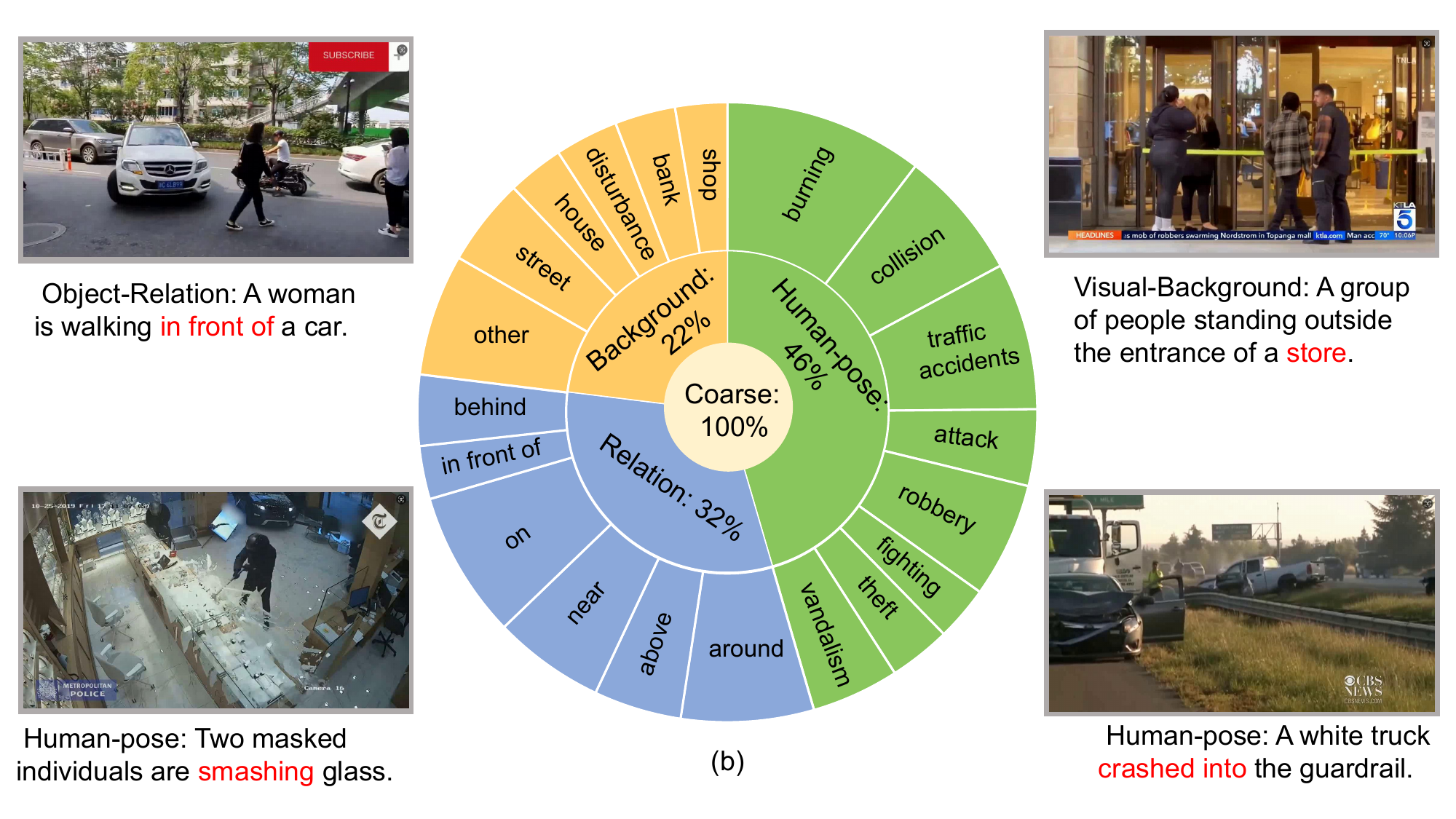}
  \caption{The statistics of physical-world information for (a) the UCF-C 
  instructions dataset and (b) the CUVA instructions. These pie charts illustrate the imbalanced distribution of these information, with coarse-grained information and human-pose comprising the largest and second-largest segments. The surrounding images provide examples corresponding to each fine-grained physical-world information. The proportions are computed based on the frequency of each physical-world descriptions in the dataset, with segment size reflecting its relative occurrence.}
  \label{fig1:dataset}
\end{figure*}

\section{ statistical distribution of our dataset}
Figure \ref{fig1:dataset} presents the statistical distribution of physical-world information in the UCF-C instructions and CUVA instructions datasets across three types of physical-world information: human pose, visual background, and object relations.
Among them, human pose constitutes the largest proportion, accounting for 41\% and 46\% in the UCF-C instructions and CUVA instructions, respectively. Background information follows, with shares of 31\% and 22\%, while object relations account for 28\% and 32\%.
This imbalance in the distribution of physical-world information may cause the model to overemphasize high-proportion categories (e.g., human pose and coarse-grained experts) while neglecting those with lower proportions (e.g., object relations).

\section{Further Convergence Analysis of UPRM}
\begin{figure} 
    \centering
    \includegraphics[width=0.235\textwidth]{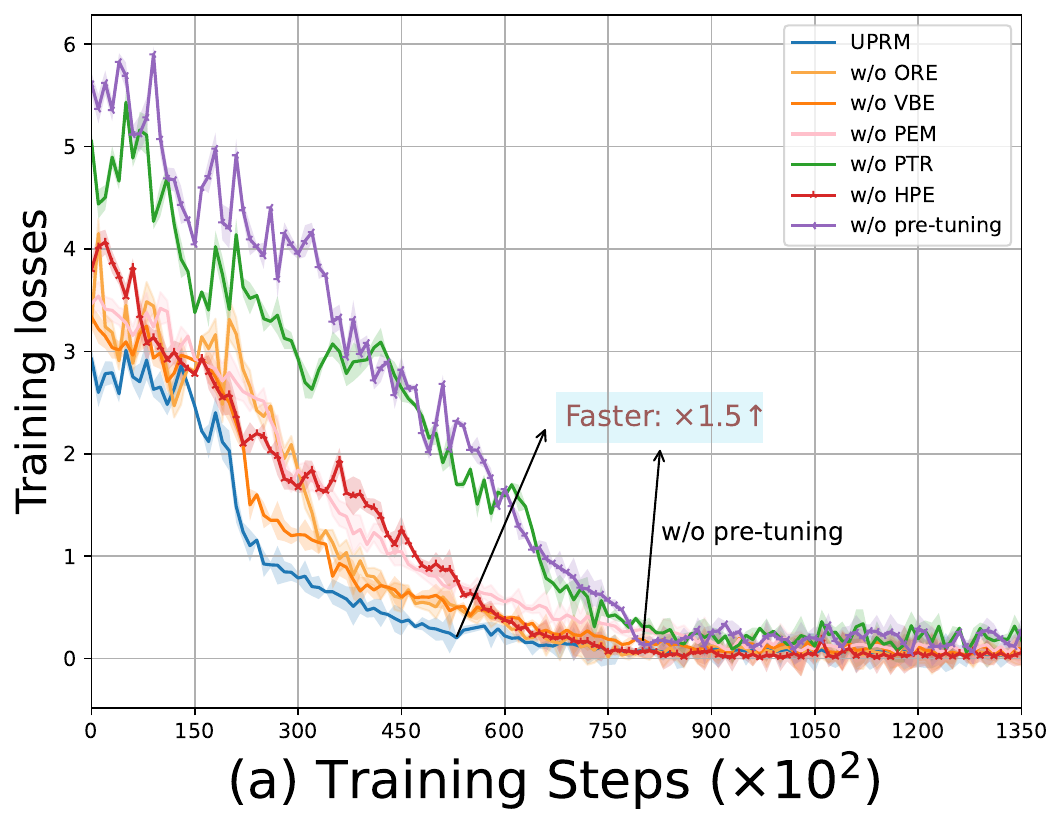}
      \hfill%
    \includegraphics[width=0.235\textwidth]{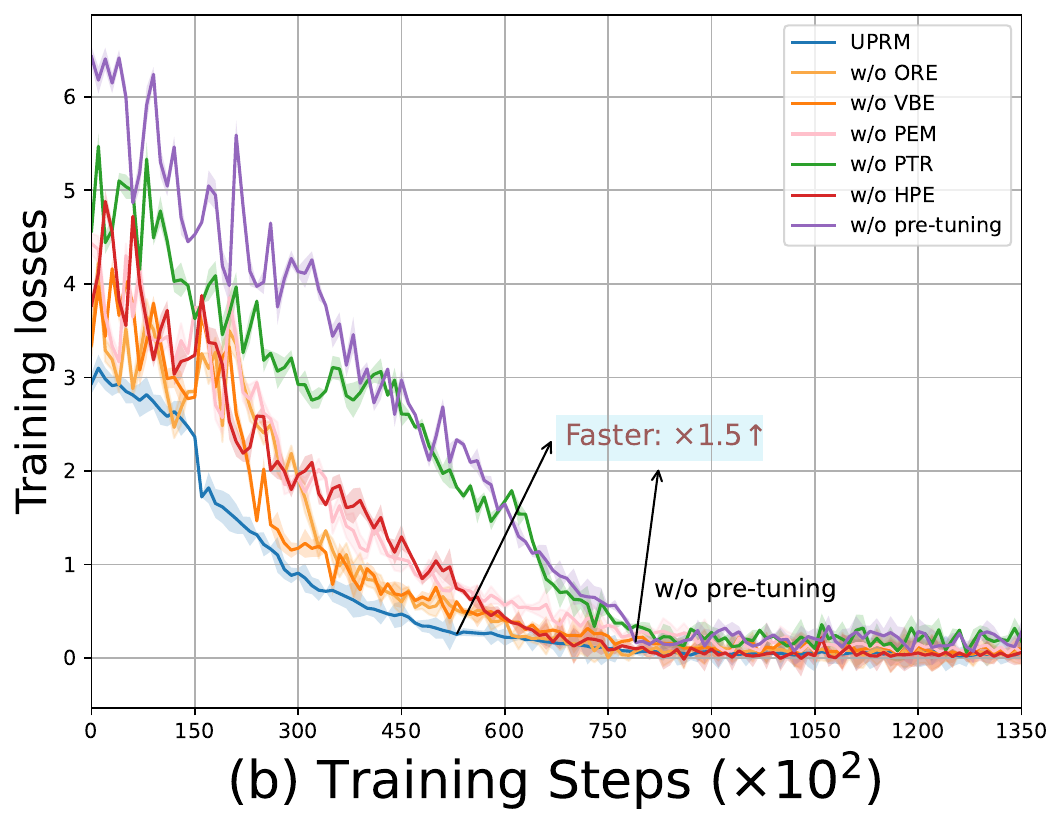}
    
    \caption{\textbf{Convergence Analysis.} Investigation of the effect of component removal within the UPRM model on (a) CUVA and (b) UCF-C instructions datasets, illustrating the differences in training losses when specific components are excluded. ``Faster" refers to how many times faster our method's convergence point is compared to the slowest baseline.}
 
   \label{fig:loss2}
\end{figure}

In Figure \ref{fig:loss2}, we present a detailed comparison of the convergence speeds of UPRM and its ablated variants across various training steps, thereby illustrating how each model's performance evolves throughout the training process. The following observations can be made from the figure. 
\textbf{1)} In the ablation study, models without components like HPE, ORE, VBE, and PEM exhibit slower convergence or higher final loss values. Notably, the model without HPE converges slower initially but matches other models later, while the model without PEM converges faster but has a higher final loss. \textbf{This demonstrates that incorporating physical-world information, especially with PEM, accelerates convergence, emphasizing the importance of physical-world understanding in the Inter-Aila task.} \textbf{2)} The model without pre-tuning converges significantly slower. \textbf{This highlights the necessity of pre-tuning to speed up convergence and enhance final performance, suggesting that better scene understanding datasets are crucial before fine-tuning UPRM.} 

\begin{figure}[t!]
    \centering
    \hspace{-0.5cm}
    \includegraphics[width=0.5\textwidth, scale=0.39]{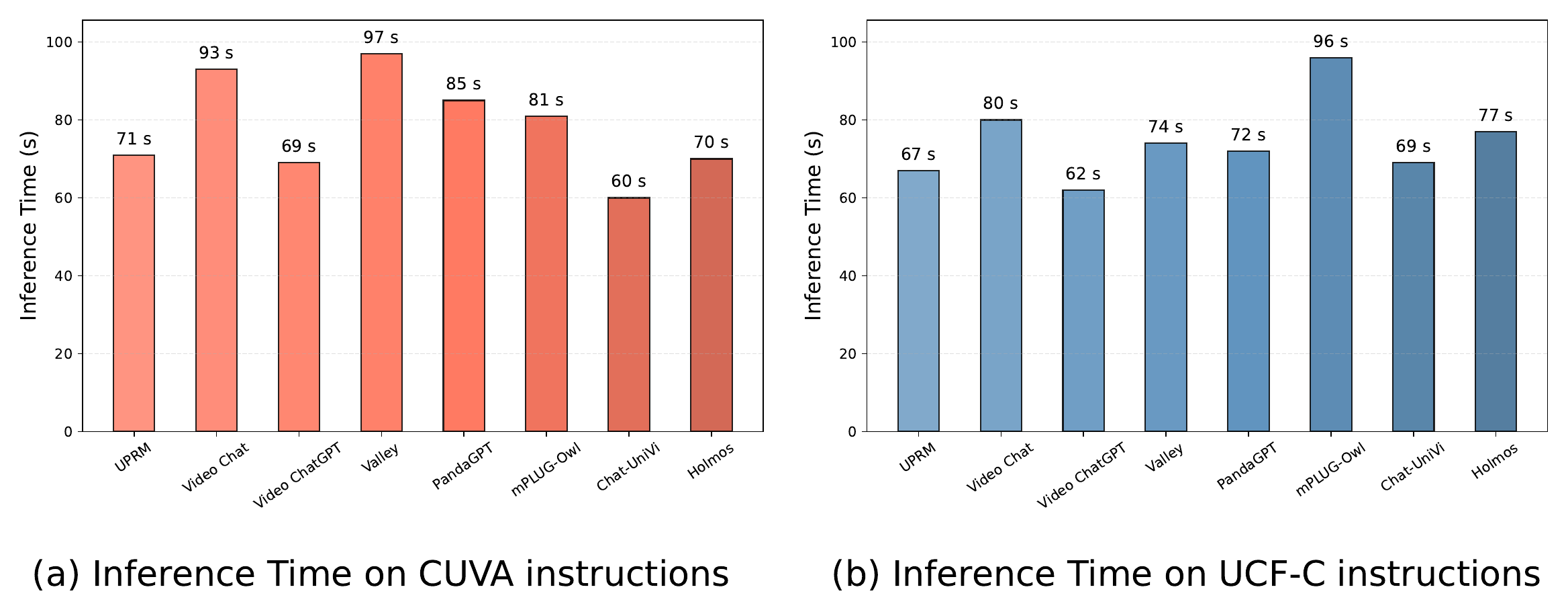}
    \caption{Comparison of the average inference time for a one-minute
video between UPRM and other Video-LLMs.}
    \label{fig:bar666}
\end{figure}

Furthermore, we conduct a comprehensive analysis comparing the inference speeds across different models. Figure \ref{fig:bar666} shows that while UPRM achieves higher accuracy, it slightly lags behind other models in inference speed. \textbf{This indicates that the accuracy gain comes at the cost of speed, which should be considered for real-time applications.}

% #表2
\begin{table*}[]
\setlength{\abovecaptionskip}{-0.1 ex}
\setlength{\belowcaptionskip}{-0.5 ex}
\caption{Comparison of our UPRM with several advanced Video-LLMs on different physical-world scenes (a total of 13 scenes) from the UCF-C instructions dataset, using FNRs.}
\setlength{\belowcaptionskip}{-5 ex}
\label{tab:33}
\resizebox{\linewidth}{!}{
\begin{tabular}{c|ccccccccccccc}
% \toprule[1.2pt]
\specialrule{1.2pt}{0pt}{0pt}

Approaches             & Abuse & \multicolumn{1}{l}{Arrest} & \multicolumn{1}{l}{Arson} & \multicolumn{1}{l}{Assault} & \multicolumn{1}{l}{Accidenct} & \multicolumn{1}{l}{Burglary} & \multicolumn{1}{l}{Explosion} & Fighting & \multicolumn{1}{l}{Robbery} & \multicolumn{1}{l}{Shooting} & \multicolumn{1}{l}{Stealing} & \multicolumn{1}{l}{Shoplifting} & \multicolumn{1}{l}{Vandalism} \\ \hline
VadCLIP\textsuperscript{†}        & 100 & 89.93                      & 68.79                     & 92.73                       & 95.05                         & 48.01                         & 72.44                         & 78.71    & 53.38                       & 68.42                        & 48.31                        & 49.28                           & 88.31                 \\
BiConvLSTM\textsuperscript{†}        & 100 & 85.07                      & 68.57                     & 95.49                       & 92.08                         & 77.36                         & 70.21                         & 91.56    & 86.47                       & 77.99                        & 88.41                        & 92.09                           & 68.83                 \\
X3D\textsuperscript{†}        & 100 & 69.03                      & 65.36                     & 95.49                       & 81.19                         & 67.03                         & 89.72                         & 85.71    & 92.48                       & 61.84                        & 79.71                       & 96.4                           & 51.95                 \\
CLIP-TSA\textsuperscript{†}        & 100 &46.27	&72.14	&61.81	&83.17	&61.78	&86.52	&71.43	&77.44	&56.82	&48.31	&73.38	&53.25
                 \\
Video Chat        & 87.5 & 42.16                      & 81.07                     & 78.47                       & 47.52                         & 56.34                         & 74.11                         & 49.91    & 68.35                       & 54.6                        & 64.25                        & 71.58                           & 66.23                          \\
Video ChatGPT     & 93.75 & 70.7                      & 76.62                     & 94.45                       & 83.67                         & 78.45                         & 78.9                         & 71.76    & 65.42                       & 70.34                        & 79.95                        & 87.59                           & 83.12                          \\
Valley             & 69.5 & 49.63                      & 52.86                     & 79.51                       & 84.16                         & 55.62                         & 50.35                         & 50    & 100                       & 92.76                        & 74.88                        & 88.49                           & 72.73                          \\
PandaGPT         & 72.5 & 70.52                     & 66.07                    & 72.57                       & 66.68                         & 67.94                         & 68.62                        & 69.81    & 68.05                       & 68.39                       & 70.78                        & 67.99                           & 66.24                          \\
mPLUG-Owl         & 75 & 50.37                      & 55                    & 68.61                       & 59.41                         & 55.25                         & 58.87                         & 56.49    & 61.63                       & 55.43                        & 43                       & 56.47                           & 66.23                          \\
Chat-UniVi       & 100 & 76.87                      & 87.5                     & 85.42                       & 94.06                         & 74.64                         & 87.59                         & 85.06    & 72.18                      & 81.06                        & 45.89                        & 97.48                          & 80.52                         \\
Video-LLaVA       & 100 & 69.77                      & 67.38                    & 91.78                       & 86.14                         & 70.41                        & 78.96                         & 77.71    & 85.96                       & 72.24                        & 68.28                        & 89.69                           & 64.94                          \\
Holmos       & 75 & \textbf{38.06}                      & 27.14                    & 35.35                       & 46.73                         & 33.88                        & \textbf{37.02}                         & 32.99    & 66.17                       & 32.84                        & \textbf{22.22}                        & 72.3                           & 28.57                          \\
Hawkeye       & 62.5 & 40.13                      & 23.13                    & 45.68                       & \textbf{42.32}                         & 35.13                        & 38.91                         & 41.15    & 59.37                       & 35.88                           & 35.46                        & \textbf{51.74}                           & 24.64                          \\
\hline
\rowcolor{lightpink}\textbf{UPRM}
           & \textcolor{blue}{61.54} & \textcolor{blue}{50.51}                      & \textcolor{blue}{19.93}                     & \textcolor{blue}{28.76}                       & \textcolor{blue}{49.65}                      & \textcolor{blue}{32.84}                         & \textcolor{blue}{48.81}                         & \textcolor{blue}{28.39}    & \textcolor{blue}{47.41}                       & \textcolor{blue}{31.64}                        & \textcolor{blue}{37.55}                        & \textcolor{blue}{64.72}                           & \textcolor{blue}{20.99}                          \\
% \bottomrule[1.2pt]
\specialrule{1.2pt}{0pt}{0pt}
\end{tabular}
}

\end{table*}

% #表3
\begin{table*}[]
\setlength{\abovecaptionskip}{-0.1 ex}
\setlength{\belowcaptionskip}{-0.5 ex}
\caption{Comparison of our UPRM with several advanced Video-LLMs on different physical-world scenes (a total of 11 scenes) from the CUVA instructions dataset, using FNRs.
}
\setlength{\belowcaptionskip}{-5 ex}
\label{tab:44}
\resizebox{\linewidth}{!}{
\begin{tabular}{c|ccccccccccccc}
% \toprule[1.2pt]
\specialrule{1.2pt}{0pt}{0pt}

Approaches             & Fire & \multicolumn{1}{l}{ Violations} & \multicolumn{1}{l}{Pedestrain } & \multicolumn{1}{l}{Animals} & \multicolumn{1}{l}{Water Incident} & \multicolumn{1}{l}{Vandalism} & \multicolumn{1}{l}{Traffic Accidents} & Robbery & \multicolumn{1}{l}{Theft} & \multicolumn{1}{l}{Fighting} & \multicolumn{1}{l}{Burning}  \\ \hline
VadCLIP\textsuperscript{†}      &66.99 &73.81 &55.36 &63.34 &59.53 &58.64 &66.03 &61.33 &68.57 &57 &65.39
\\
BiConvLSTM\textsuperscript{†}    &64.42 &71.34 &50.92 &64.07 &56.91 &57.94 &63.72 &68.74 &72.42 &51.35 &64.93                                              \\
X3D\textsuperscript{†}       & 41.58 &21.56 &43.42 &64.29 &42.94 &35.7 &40.23 &17.22 &38.55 &36.85 &48.81                                             \\
CLIP-TSA\textsuperscript{†}        &59.76  &58.29  &58.34 &59.24 &58.11 &60.02 &59.87 &60.45 &56.93 &55.62 &57.88                                          \\
Video Chat       &47.73 &21.9 &47.29 &30.78 &20.61 &27.26 &20.19 &44.81 &63.06 &50.34 &26.84                                                 \\
Video ChatGPT     & 20.94 &46.15 &36.09 &51.68 &40.27 &38.3 &35.19 &45.67  &38.23  &32.36 &36.38                                           \\
Valley             & 58.93 &59.57 &57.81 &58.8 &59.53 &57.67 &60.66 &60.68 &57.26 &56.97 &58.6                                               \\
PandaGPT       &21.54  &37.92    &12.14  &27.89  &12.55  &45.47  &17.92  &10.5  &23.39  &19.44  &27.05                      \\
mPLUG-Owl         & 52.93  &46.93  &45.55  &33.39  &43.85  &42.51  &  40.03  &51.16  &52.42  &33.93  &41.95                    \\
Chat-UniVi       & 21.75  &46.64  &35.61  &51.84  &39.55  &37.09  &35.48  &46.38  &35.4  &31.94  &36.45                                           \\
Video-LLaVA      & 	17.43  &70.56  &6.28  &13.41  &5.28  &60.58  &39.31  &11.24  &54.47  &30.79  &18.07                                              \\ 
Holmos    &  11.86  & 27.62  & 13.15 & 24.7 &12.79 &28.3 &21.42 &7.14 &46.61 &21.69   &14.29                                        \\ 
Hawkeye     & 7.13 &23.71  & 8.64  &26.13 &3.03 &30.6 &15.36 &6.94 &37.18 &13.64 &10.89                                            \\ 
\hline
\rowcolor{lightpink}\textbf{UPRM} & \textcolor{blue}{5.93} & 
\textcolor{blue}{18.01} & \textcolor{blue}{3.69} & \textcolor{blue}{20.68} & \textcolor{blue}{2.02} & \textcolor{blue}{26.1} & \textcolor{blue}{8.19} & \textcolor{blue}{4.64} & \textcolor{blue}{28.06} & \textcolor{blue}{8.76} & \textcolor{blue}{3.4}                                      \\
% \bottomrule[1.2pt]
\specialrule{1.2pt}{0pt}{0pt}
\end{tabular}
}

\end{table*}

\section{Real-world Robustness Analysis for UPRM }

We focus on the performance of various SVU models across different physical-world scenes, with particular emphasis on the superior performance of the UPRM  model in terms of False Negative Rates (FNRs). The relevant results are presented in Tables \ref{tab:33} and \ref{tab:44}.

UPRM  consistently outperforms both non-LLM anomaly detection approaches and Video-LLMs in most scenarios, particularly in reducing FNRs.
As shown in Table \ref{tab:33}, UPRM consistently outperforms most models, including non-LLM approaches and Video-LLMs. For example, in the Fighting and Assault categories, UPRM surpasses the second-best experimental results, with FNRs of 3.6 and 6.59, respectively. However, despite surpassing most models, UPRM's performance in the Arrest, Accident, Explosion, Shoplifting and Stealing categories still falls behind the best experimental results, highlighting an area for improvement.
Table \ref{tab:44} shows UPRM’s dominance across the CUVA instructions dataset, where it outperforms all other models. For instance, UPRM achieves FNRs of 18.01 and 3.4 in Violations and Burning, surpassing the best model by 5.7 and 7.49. 
\textbf{These results underline UPRM’s effectiveness in diverse physical-world scenes and emphasize the need to leverage specific expert strengths for the DeepSVU task.} The model’s dominance is especially crucial in detecting subtle anomalies in complex scenarios like Vandalism and Robbery, where precise detection is vital for effective surveillance and response.

%% file: sample-sigconf.bbl
\begin{thebibliography}{49}

%%% the Web address.

\ifx \showCODEN    \undefined \def \showCODEN     #1{\unskip}     \fi
\ifx \showISBNx    \undefined \def \showISBNx     #1{\unskip}     \fi
\ifx \showISBNxiii \undefined \def \showISBNxiii  #1{\unskip}     \fi
\ifx \showISSN     \undefined \def \showISSN      #1{\unskip}     \fi
\ifx \showLCCN     \undefined \def \showLCCN      #1{\unskip}     \fi
\ifx \shownote     \undefined \def \shownote      #1{#1}          \fi
\ifx \showarticletitle \undefined \def \showarticletitle #1{#1}   \fi
\ifx \showURL      \undefined \def \showURL       {\relax}        \fi
% invisible to TeX
\providecommand\bibfield[2]{#2}
\providecommand\bibinfo[2]{#2}
\providecommand\natexlab[1]{#1}
\providecommand\showeprint[2][]{arXiv:#2}

\bibitem[Ba et~al\mbox{.}(2016)]%
        {layernorm}
\bibfield{author}{\bibinfo{person}{Jimmy~Lei Ba}, \bibinfo{person}{Jamie~Ryan Kiros}, {and} \bibinfo{person}{Geoffrey~E Hinton}.} \bibinfo{year}{2016}\natexlab{}.
\newblock \showarticletitle{Layer normalization}.
\newblock \bibinfo{journal}{\emph{arXiv preprint arXiv:1607.06450}} (\bibinfo{year}{2016}).
\newblock


\bibitem[Chen et~al\mbox{.}(2024)]%
        {chen2024vast}
\bibfield{author}{\bibinfo{person}{Sihan Chen}, \bibinfo{person}{Handong Li}, \bibinfo{person}{Qunbo Wang}, \bibinfo{person}{Zijia Zhao}, \bibinfo{person}{Mingzhen Sun}, \bibinfo{person}{Xinxin Zhu}, {and} \bibinfo{person}{Jing Liu}.} \bibinfo{year}{2024}\natexlab{}.
\newblock \showarticletitle{Vast: A vision-audio-subtitle-text omni-modality foundation model and dataset}.
\newblock \bibinfo{journal}{\emph{Proceedings of NeurIPS}}  \bibinfo{volume}{36} (\bibinfo{year}{2024}).
\newblock


\bibitem[Cheng et~al\mbox{.}(2020)]%
        {cheng2020higherhrnet}
\bibfield{author}{\bibinfo{person}{Bowen Cheng}, \bibinfo{person}{Bin Xiao}, \bibinfo{person}{Jingdong Wang}, \bibinfo{person}{Honghui Shi}, \bibinfo{person}{Thomas~S Huang}, {and} \bibinfo{person}{Lei Zhang}.} \bibinfo{year}{2020}\natexlab{}.
\newblock \showarticletitle{Higherhrnet: Scale-aware representation learning for bottom-up human pose estimation}. In \bibinfo{booktitle}{\emph{Proceedings of CVPR}}. \bibinfo{pages}{5386--5395}.
\newblock


\bibitem[Cong et~al\mbox{.}(2023)]%
        {reltr}
\bibfield{author}{\bibinfo{person}{Yuren Cong}, \bibinfo{person}{Michael~Ying Yang}, {and} \bibinfo{person}{Bodo Rosenhahn}.} \bibinfo{year}{2023}\natexlab{}.
\newblock \showarticletitle{Reltr: Relation transformer for scene graph generation}.
\newblock \bibinfo{journal}{\emph{IEEE Transactions on Pattern Analysis and Machine Intelligence}} (\bibinfo{year}{2023}).
\newblock


\bibitem[Du et~al\mbox{.}(2024)]%
        {cuva}
\bibfield{author}{\bibinfo{person}{Hang Du}, \bibinfo{person}{Sicheng Zhang}, \bibinfo{person}{Binzhu Xie}, \bibinfo{person}{Guoshun Nan}, \bibinfo{person}{Jiayang Zhang}, \bibinfo{person}{Junrui Xu}, \bibinfo{person}{Hangyu Liu}, \bibinfo{person}{Sicong Leng}, \bibinfo{person}{Jiangming Liu}, \bibinfo{person}{Hehe Fan}, {et~al\mbox{.}}} \bibinfo{year}{2024}\natexlab{}.
\newblock \showarticletitle{Uncovering What Why and How: A Comprehensive Benchmark for Causation Understanding of Video Anomaly}. In \bibinfo{booktitle}{\emph{Proceedings of CVPR}}. \bibinfo{pages}{18793--18803}.
\newblock


\bibitem[Edstedt et~al\mbox{.}(2022)]%
        {edstedt2022vidharm}
\bibfield{author}{\bibinfo{person}{Johan Edstedt}, \bibinfo{person}{Amanda Berg}, \bibinfo{person}{Michael Felsberg}, \bibinfo{person}{Johan Karlsson}, \bibinfo{person}{Francisca Benavente}, \bibinfo{person}{Anette Novak}, {and} \bibinfo{person}{Gustav~Grund Pihlgren}.} \bibinfo{year}{2022}\natexlab{}.
\newblock \showarticletitle{Vidharm: A clip based dataset for harmful content detection}. In \bibinfo{booktitle}{\emph{2022 26th International Conference on Pattern Recognition (ICPR)}}. IEEE, \bibinfo{pages}{1543--1549}.
\newblock


\bibitem[Guo et~al\mbox{.}(2022)]%
        {guo2022generating}
\bibfield{author}{\bibinfo{person}{Chuan Guo}, \bibinfo{person}{Shihao Zou}, \bibinfo{person}{Xinxin Zuo}, \bibinfo{person}{Sen Wang}, \bibinfo{person}{Wei Ji}, \bibinfo{person}{Xingyu Li}, {and} \bibinfo{person}{Li Cheng}.} \bibinfo{year}{2022}\natexlab{}.
\newblock \showarticletitle{Generating diverse and natural 3d human motions from text}. In \bibinfo{booktitle}{\emph{Proceedings of the CVPR}}. \bibinfo{pages}{5152--5161}.
\newblock


\bibitem[Han et~al\mbox{.}(2024)]%
        {onellm}
\bibfield{author}{\bibinfo{person}{Jiaming Han}, \bibinfo{person}{Kaixiong Gong}, \bibinfo{person}{Yiyuan Zhang}, \bibinfo{person}{Jiaqi Wang}, \bibinfo{person}{Kaipeng Zhang}, \bibinfo{person}{Dahua Lin}, \bibinfo{person}{Yu Qiao}, \bibinfo{person}{Peng Gao}, {and} \bibinfo{person}{Xiangyu Yue}.} \bibinfo{year}{2024}\natexlab{}.
\newblock \showarticletitle{OneLLM: One Framework to Align All Modalities with Language}.
\newblock  (\bibinfo{year}{2024}), \bibinfo{pages}{26584--26595}.
\newblock


\bibitem[Hanson et~al\mbox{.}(2018)]%
        {BiConvLSTM}
\bibfield{author}{\bibinfo{person}{Alex Hanson}, \bibinfo{person}{Koutilya Pnvr}, \bibinfo{person}{Sanjukta Krishnagopal}, {and} \bibinfo{person}{Larry Davis}.} \bibinfo{year}{2018}\natexlab{}.
\newblock \showarticletitle{Bidirectional convolutional lstm for the detection of violence in videos}. In \bibinfo{booktitle}{\emph{Proceedings of ECCV workshops}}. \bibinfo{pages}{0--0}.
\newblock


\bibitem[Hasan et~al\mbox{.}(2016)]%
        {hasan2016learning}
\bibfield{author}{\bibinfo{person}{Mahmudul Hasan}, \bibinfo{person}{Jonghyun Choi}, \bibinfo{person}{Jan Neumann}, \bibinfo{person}{Amit~K Roy-Chowdhury}, {and} \bibinfo{person}{Larry~S Davis}.} \bibinfo{year}{2016}\natexlab{}.
\newblock \showarticletitle{Learning temporal regularity in video sequences}. In \bibinfo{booktitle}{\emph{Proceedings of CVPR}}. \bibinfo{pages}{733--742}.
\newblock


\bibitem[Hu et~al\mbox{.}(2022)]%
        {lora}
\bibfield{author}{\bibinfo{person}{Edward~J. Hu}, \bibinfo{person}{Yelong Shen}, \bibinfo{person}{Phillip Wallis}, \bibinfo{person}{Zeyuan Allen{-}Zhu}, \bibinfo{person}{Yuanzhi Li}, \bibinfo{person}{Shean Wang}, \bibinfo{person}{Lu Wang}, {and} \bibinfo{person}{Weizhu Chen}.} \bibinfo{year}{2022}\natexlab{}.
\newblock \showarticletitle{LoRA: Low-Rank Adaptation of Large Language Models}. In \bibinfo{booktitle}{\emph{Proceedings of {ICLR} 2022}}.
\newblock


\bibitem[Huang et~al\mbox{.}(2025)]%
        {huang2025exvad}
\bibfield{author}{\bibinfo{person}{C. Huang}, \bibinfo{person}{Y. Shi}, \bibinfo{person}{J. Wen}, {et~al\mbox{.}}} \bibinfo{year}{2025}\natexlab{}.
\newblock \showarticletitle{Ex-VAD: Explainable Fine-grained Video Anomaly Detection Based on Visual-Language Models}. In \bibinfo{booktitle}{\emph{Proceedings of the Forty-second International Conference on Machine Learning}}.
\newblock


\bibitem[Jin et~al\mbox{.}(2022)]%
        {jin2022anomaly}
\bibfield{author}{\bibinfo{person}{P. Jin}, \bibinfo{person}{L. Mou}, \bibinfo{person}{G.~S. Xia}, {et~al\mbox{.}}} \bibinfo{year}{2022}\natexlab{}.
\newblock \showarticletitle{Anomaly detection in aerial videos with transformers}.
\newblock \bibinfo{journal}{\emph{IEEE Transactions on Geoscience and Remote Sensing}}  \bibinfo{volume}{60} (\bibinfo{year}{2022}), \bibinfo{pages}{1--13}.
\newblock


\bibitem[Jin et~al\mbox{.}(2023)]%
        {chatunivi}
\bibfield{author}{\bibinfo{person}{Peng Jin}, \bibinfo{person}{Ryuichi Takanobu}, \bibinfo{person}{Caiwan Zhang}, \bibinfo{person}{Xiaochun Cao}, {and} \bibinfo{person}{Li Yuan}.} \bibinfo{year}{2023}\natexlab{}.
\newblock \showarticletitle{Chat-UniVi: Unified Visual Representation Empowers Large Language Models with Image and Video Understanding}.
\newblock \bibinfo{journal}{\emph{CoRR}}  \bibinfo{volume}{abs/2311.08046} (\bibinfo{year}{2023}).
\newblock


\bibitem[Joo et~al\mbox{.}(2023)]%
        {clip-tsa}
\bibfield{author}{\bibinfo{person}{Hyekang~Kevin Joo}, \bibinfo{person}{Khoa Vo}, \bibinfo{person}{Kashu Yamazaki}, {and} \bibinfo{person}{Ngan Le}.} \bibinfo{year}{2023}\natexlab{}.
\newblock \showarticletitle{Clip-tsa: Clip-assisted temporal self-attention for weakly-supervised video anomaly detection}. In \bibinfo{booktitle}{\emph{Proceedings of ICIP}}. IEEE, \bibinfo{pages}{3230--3234}.
\newblock


\bibitem[Kirillov et~al\mbox{.}(2023a)]%
        {kirillov2023segment}
\bibfield{author}{\bibinfo{person}{Alexander Kirillov}, \bibinfo{person}{Eric Mintun}, \bibinfo{person}{Nikhila Ravi}, \bibinfo{person}{Hanzi Mao}, \bibinfo{person}{Chloe Rolland}, \bibinfo{person}{Laura Gustafson}, \bibinfo{person}{Tete Xiao}, \bibinfo{person}{Spencer Whitehead}, \bibinfo{person}{Alexander~C Berg}, \bibinfo{person}{Wan-Yen Lo}, {et~al\mbox{.}}} \bibinfo{year}{2023}\natexlab{a}.
\newblock \showarticletitle{Segment anything}. In \bibinfo{booktitle}{\emph{Proceedings of CVPR}}. \bibinfo{pages}{4015--4026}.
\newblock


\bibitem[Kirillov et~al\mbox{.}(2023b)]%
        {sam}
\bibfield{author}{\bibinfo{person}{Alexander Kirillov}, \bibinfo{person}{Eric Mintun}, \bibinfo{person}{Nikhila Ravi}, \bibinfo{person}{Hanzi Mao}, \bibinfo{person}{Chloe Rolland}, \bibinfo{person}{Laura Gustafson}, \bibinfo{person}{Tete Xiao}, \bibinfo{person}{Spencer Whitehead}, \bibinfo{person}{Alexander~C Berg}, \bibinfo{person}{Wan-Yen Lo}, {et~al\mbox{.}}} \bibinfo{year}{2023}\natexlab{b}.
\newblock \showarticletitle{Segment anything}. In \bibinfo{booktitle}{\emph{Proceedings of the IEEE/CVF International Conference on Computer Vision}}. \bibinfo{pages}{4015--4026}.
\newblock


\bibitem[Li et~al\mbox{.}(2017)]%
        {RSI-CB}
\bibfield{author}{\bibinfo{person}{Haifeng Li}, \bibinfo{person}{Xin Dou}, \bibinfo{person}{Chao Tao}, \bibinfo{person}{Zhixiang Hou}, \bibinfo{person}{Jie Chen}, \bibinfo{person}{Jian Peng}, \bibinfo{person}{Min Deng}, {and} \bibinfo{person}{Ling Zhao}.} \bibinfo{year}{2017}\natexlab{}.
\newblock \showarticletitle{RSI-CB: A large scale remote sensing image classification benchmark via crowdsource data}.
\newblock \bibinfo{journal}{\emph{arXiv preprint arXiv:1705.10450}} (\bibinfo{year}{2017}).
\newblock


\bibitem[Li et~al\mbox{.}(2024a)]%
        {li2024cumo}
\bibfield{author}{\bibinfo{person}{Jiachen Li}, \bibinfo{person}{Xinyao Wang}, \bibinfo{person}{Sijie Zhu}, \bibinfo{person}{Chia-Wen Kuo}, \bibinfo{person}{Lu Xu}, \bibinfo{person}{Fan Chen}, \bibinfo{person}{Jitesh Jain}, \bibinfo{person}{Humphrey Shi}, {and} \bibinfo{person}{Longyin Wen}.} \bibinfo{year}{2024}\natexlab{a}.
\newblock \showarticletitle{Cumo: Scaling multimodal llm with co-upcycled mixture-of-experts}.
\newblock \bibinfo{journal}{\emph{arXiv preprint arXiv:2405.05949}} (\bibinfo{year}{2024}).
\newblock


\bibitem[Li et~al\mbox{.}(2023)]%
        {videochat}
\bibfield{author}{\bibinfo{person}{Kunchang Li}, \bibinfo{person}{Yinan He}, \bibinfo{person}{Yi Wang}, \bibinfo{person}{Yizhuo Li}, \bibinfo{person}{Wenhai Wang}, \bibinfo{person}{Ping Luo}, \bibinfo{person}{Yali Wang}, \bibinfo{person}{Limin Wang}, {and} \bibinfo{person}{Yu Qiao}.} \bibinfo{year}{2023}\natexlab{}.
\newblock \showarticletitle{VideoChat: Chat-Centric Video Understanding}.
\newblock \bibinfo{journal}{\emph{CoRR}}  \bibinfo{volume}{abs/2305.06355} (\bibinfo{year}{2023}).
\newblock


\bibitem[Li et~al\mbox{.}({[n.\,d.]})]%
        {ligeoreasoner}
\bibfield{author}{\bibinfo{person}{Ling Li}, \bibinfo{person}{Yu Ye}, \bibinfo{person}{Bingchuan Jiang}, {and} \bibinfo{person}{Wei Zeng}.} \bibinfo{year}{[n.\,d.]}\natexlab{}.
\newblock \showarticletitle{GeoReasoner: Geo-localization with Reasoning in Street Views using a Large Vision-Language Model}. In \bibinfo{booktitle}{\emph{Proceedings of {ICML} 2024}}.
\newblock


\bibitem[Li et~al\mbox{.}(2024b)]%
        {li20243dmit}
\bibfield{author}{\bibinfo{person}{Zeju Li}, \bibinfo{person}{Chao Zhang}, \bibinfo{person}{Xiaoyan Wang}, \bibinfo{person}{Ruilong Ren}, \bibinfo{person}{Yifan Xu}, \bibinfo{person}{Ruifei Ma}, {and} \bibinfo{person}{Xiangde Liu}.} \bibinfo{year}{2024}\natexlab{b}.
\newblock \showarticletitle{3dmit: 3d multi-modal instruction tuning for scene understanding}.
\newblock \bibinfo{journal}{\emph{arXiv preprint arXiv:2401.03201}} (\bibinfo{year}{2024}).
\newblock


\bibitem[Lin et~al\mbox{.}(2023)]%
        {lin2023video}
\bibfield{author}{\bibinfo{person}{Bin Lin}, \bibinfo{person}{Bin Zhu}, \bibinfo{person}{Yang Ye}, \bibinfo{person}{Munan Ning}, \bibinfo{person}{Peng Jin}, {and} \bibinfo{person}{Li Yuan}.} \bibinfo{year}{2023}\natexlab{}.
\newblock \showarticletitle{Video-llava: Learning united visual representation by alignment before projection}.
\newblock  (\bibinfo{year}{2023}).
\newblock


\bibitem[Lin et~al\mbox{.}(2014)]%
        {lin2014microsoft}
\bibfield{author}{\bibinfo{person}{Tsung-Yi Lin}, \bibinfo{person}{Michael Maire}, \bibinfo{person}{Serge Belongie}, \bibinfo{person}{James Hays}, \bibinfo{person}{Pietro Perona}, \bibinfo{person}{Deva Ramanan}, \bibinfo{person}{Piotr Doll{\'a}r}, {and} \bibinfo{person}{C~Lawrence Zitnick}.} \bibinfo{year}{2014}\natexlab{}.
\newblock \showarticletitle{Microsoft coco: Common objects in context}. In \bibinfo{booktitle}{\emph{Proceedings of ECCV}}. Springer, \bibinfo{pages}{740--755}.
\newblock


\bibitem[Luo et~al\mbox{.}(2025)]%
        {luo2025omnisila}
\bibfield{author}{\bibinfo{person}{J. Luo}, \bibinfo{person}{J. Wang}, \bibinfo{person}{J. Ma}, {et~al\mbox{.}}} \bibinfo{year}{2025}\natexlab{}.
\newblock \showarticletitle{Omni-SILA: Towards Omni-scene Driven Visual Sentiment Identifying, Locating and Attributing in Videos}. In \bibinfo{booktitle}{\emph{Proceedings of the ACM on Web Conference 2025}}. \bibinfo{pages}{188--197}.
\newblock


\bibitem[Luo et~al\mbox{.}(2023)]%
        {valley}
\bibfield{author}{\bibinfo{person}{Ruipu Luo}, \bibinfo{person}{Ziwang Zhao}, \bibinfo{person}{Min Yang}, \bibinfo{person}{Junwei Dong}, \bibinfo{person}{Minghui Qiu}, \bibinfo{person}{Pengcheng Lu}, \bibinfo{person}{Tao Wang}, {and} \bibinfo{person}{Zhongyu Wei}.} \bibinfo{year}{2023}\natexlab{}.
\newblock \showarticletitle{Valley: Video Assistant with Large Language model Enhanced abilitY}.
\newblock \bibinfo{journal}{\emph{CoRR}}  \bibinfo{volume}{abs/2306.07207} (\bibinfo{year}{2023}).
\newblock


\bibitem[Ma et~al\mbox{.}(2025)]%
        {ma2025sherlock}
\bibfield{author}{\bibinfo{person}{Junxiao Ma}, \bibinfo{person}{Jingjing Wang}, \bibinfo{person}{Jiamin Luo}, \bibinfo{person}{Peiying Yu}, {and} \bibinfo{person}{Guodong Zhou}.} \bibinfo{year}{2025}\natexlab{}.
\newblock \showarticletitle{Sherlock: Towards Multi-scene Video Abnormal Event Extraction and Localization via a Global-local Spatial-sensitive LLM}. In \bibinfo{booktitle}{\emph{Proceedings of the ACM on Web Conference 2025}}. \bibinfo{pages}{4004--4013}.
\newblock


\bibitem[Mao et~al\mbox{.}(2021)]%
        {mao2021multi}
\bibfield{author}{\bibinfo{person}{Wei Mao}, \bibinfo{person}{Miaomiao Liu}, \bibinfo{person}{Mathieu Salzmann}, {and} \bibinfo{person}{Hongdong Li}.} \bibinfo{year}{2021}\natexlab{}.
\newblock \showarticletitle{Multi-level motion attention for human motion prediction}.
\newblock \bibinfo{journal}{\emph{International journal of computer vision}} \bibinfo{volume}{129}, \bibinfo{number}{9} (\bibinfo{year}{2021}), \bibinfo{pages}{2513--2535}.
\newblock


\bibitem[Munasinghe et~al\mbox{.}(2023)]%
        {munasinghe2023pg}
\bibfield{author}{\bibinfo{person}{Shehan Munasinghe}, \bibinfo{person}{Rusiru Thushara}, \bibinfo{person}{Muhammad Maaz}, \bibinfo{person}{Hanoona~Abdul Rasheed}, \bibinfo{person}{Salman Khan}, \bibinfo{person}{Mubarak Shah}, {and} \bibinfo{person}{Fahad Khan}.} \bibinfo{year}{2023}\natexlab{}.
\newblock \showarticletitle{Pg-video-llava: Pixel grounding large video-language models}.
\newblock \bibinfo{journal}{\emph{arXiv preprint arXiv:2311.13435}} (\bibinfo{year}{2023}).
\newblock


\bibitem[Puigcerver et~al\mbox{.}(2023)]%
        {moe3}
\bibfield{author}{\bibinfo{person}{Joan Puigcerver}, \bibinfo{person}{Carlos Riquelme}, \bibinfo{person}{Basil Mustafa}, {and} \bibinfo{person}{Neil Houlsby}.} \bibinfo{year}{2023}\natexlab{}.
\newblock \showarticletitle{From sparse to soft mixtures of experts}.
\newblock \bibinfo{journal}{\emph{arXiv preprint arXiv:2308.00951}} (\bibinfo{year}{2023}).
\newblock


\bibitem[Qu et~al\mbox{.}(2023)]%
        {qu2023layoutllm}
\bibfield{author}{\bibinfo{person}{Leigang Qu}, \bibinfo{person}{Shengqiong Wu}, \bibinfo{person}{Hao Fei}, \bibinfo{person}{Liqiang Nie}, {and} \bibinfo{person}{Tat-Seng Chua}.} \bibinfo{year}{2023}\natexlab{}.
\newblock \showarticletitle{Layoutllm-t2i: Eliciting layout guidance from llm for text-to-image generation}. In \bibinfo{booktitle}{\emph{Proceedings of the 31st ACM International Conference on Multimedia}}. \bibinfo{pages}{643--654}.
\newblock


\bibitem[Raychaudhuri and Chang(2025)]%
        {raychaudhuri2025semantic}
\bibfield{author}{\bibinfo{person}{Sonia Raychaudhuri} {and} \bibinfo{person}{Angel~X Chang}.} \bibinfo{year}{2025}\natexlab{}.
\newblock \showarticletitle{Semantic Mapping in Indoor Embodied AI--A Comprehensive Survey and Future Directions}.
\newblock \bibinfo{journal}{\emph{arXiv preprint arXiv:2501.05750}} (\bibinfo{year}{2025}).
\newblock


\bibitem[Su et~al\mbox{.}(2022)]%
        {x3d}
\bibfield{author}{\bibinfo{person}{Jiayi Su}, \bibinfo{person}{Paris Her}, \bibinfo{person}{Erik Clemens}, \bibinfo{person}{Edwin Yaz}, \bibinfo{person}{Susan Schneider}, {and} \bibinfo{person}{Henry Medeiros}.} \bibinfo{year}{2022}\natexlab{}.
\newblock \showarticletitle{Violence detection using 3D convolutional neural networks}. In \bibinfo{booktitle}{\emph{Proceedings of AVSS}}. IEEE, \bibinfo{pages}{1--8}.
\newblock


\bibitem[Su et~al\mbox{.}(2023)]%
        {pandagpt}
\bibfield{author}{\bibinfo{person}{Yixuan Su}, \bibinfo{person}{Tian Lan}, \bibinfo{person}{Huayang Li}, \bibinfo{person}{Jialu Xu}, \bibinfo{person}{Yan Wang}, {and} \bibinfo{person}{Deng Cai}.} \bibinfo{year}{2023}\natexlab{}.
\newblock \showarticletitle{PandaGPT: One Model To Instruction-Follow Them All}.
\newblock \bibinfo{journal}{\emph{CoRR}}  \bibinfo{volume}{abs/2305.16355} (\bibinfo{year}{2023}).
\newblock


\bibitem[Sultani et~al\mbox{.}(2018)]%
        {ucf-crime}
\bibfield{author}{\bibinfo{person}{Waqas Sultani}, \bibinfo{person}{Chen Chen}, {and} \bibinfo{person}{Mubarak Shah}.} \bibinfo{year}{2018}\natexlab{}.
\newblock \showarticletitle{Real-world anomaly detection in surveillance videos}. In \bibinfo{booktitle}{\emph{Proceedings of CVPR}}. \bibinfo{pages}{6479--6488}.
\newblock


\bibitem[Sun and Gong(2023)]%
        {sun2023hierarchical}
\bibfield{author}{\bibinfo{person}{Shengyang Sun} {and} \bibinfo{person}{Xiaojin Gong}.} \bibinfo{year}{2023}\natexlab{}.
\newblock \showarticletitle{Hierarchical semantic contrast for scene-aware video anomaly detection}. In \bibinfo{booktitle}{\emph{Proceedings of CVPR}}. \bibinfo{pages}{22846--22856}.
\newblock


\bibitem[Tang et~al\mbox{.}(2023)]%
        {tang2023video}
\bibfield{author}{\bibinfo{person}{Yunlong Tang}, \bibinfo{person}{Jing Bi}, \bibinfo{person}{Siting Xu}, \bibinfo{person}{Luchuan Song}, \bibinfo{person}{Susan Liang}, \bibinfo{person}{Teng Wang}, \bibinfo{person}{Daoan Zhang}, \bibinfo{person}{Jie An}, \bibinfo{person}{Jingyang Lin}, \bibinfo{person}{Rongyi Zhu}, {et~al\mbox{.}}} \bibinfo{year}{2023}\natexlab{}.
\newblock \showarticletitle{Video understanding with large language models: A survey}.
\newblock \bibinfo{journal}{\emph{arXiv preprint arXiv:2312.17432}} (\bibinfo{year}{2023}).
\newblock


\bibitem[Tran et~al\mbox{.}(2015)]%
        {Tran_Bourdev_Fergus_Torresani_Paluri_2015}
\bibfield{author}{\bibinfo{person}{Du Tran}, \bibinfo{person}{Lubomir Bourdev}, \bibinfo{person}{Rob Fergus}, \bibinfo{person}{Lorenzo Torresani}, {and} \bibinfo{person}{Manohar Paluri}.} \bibinfo{year}{2015}\natexlab{}.
\newblock \showarticletitle{Learning Spatiotemporal Features with 3D Convolutional Networks}. In \bibinfo{booktitle}{\emph{Proceedings of ICCV}}.
\newblock
\href{https://doi.org/10.1109/iccv.2015.510}{doi:\nolinkurl{10.1109/iccv.2015.510}}


\bibitem[Wu et~al\mbox{.}(2024)]%
        {vadclip}
\bibfield{author}{\bibinfo{person}{Peng Wu}, \bibinfo{person}{Xuerong Zhou}, \bibinfo{person}{Guansong Pang}, \bibinfo{person}{Lingru Zhou}, \bibinfo{person}{Qingsen Yan}, \bibinfo{person}{Peng Wang}, {and} \bibinfo{person}{Yanning Zhang}.} \bibinfo{year}{2024}\natexlab{}.
\newblock \showarticletitle{Vadclip: Adapting vision-language models for weakly supervised video anomaly detection}. In \bibinfo{booktitle}{\emph{Proceedings of AAAI}}.
\newblock


\bibitem[Ye et~al\mbox{.}(2023)]%
        {ye2023mplug}
\bibfield{author}{\bibinfo{person}{Qinghao Ye}, \bibinfo{person}{Haiyang Xu}, \bibinfo{person}{Guohai Xu}, \bibinfo{person}{Jiabo Ye}, \bibinfo{person}{Ming Yan}, \bibinfo{person}{Yiyang Zhou}, \bibinfo{person}{Junyang Wang}, \bibinfo{person}{Anwen Hu}, \bibinfo{person}{Pengcheng Shi}, \bibinfo{person}{Yaya Shi}, {et~al\mbox{.}}} \bibinfo{year}{2023}\natexlab{}.
\newblock \showarticletitle{mplug-owl: Modularization empowers large language models with multimodality}.
\newblock \bibinfo{journal}{\emph{arXiv preprint arXiv:2304.14178}} (\bibinfo{year}{2023}).
\newblock


\bibitem[Yousaf and Nawaz(2022)]%
        {yousaf2022deep}
\bibfield{author}{\bibinfo{person}{Kanwal Yousaf} {and} \bibinfo{person}{Tabassam Nawaz}.} \bibinfo{year}{2022}\natexlab{}.
\newblock \showarticletitle{A deep learning-based approach for inappropriate content detection and classification of youtube videos}.
\newblock \bibinfo{journal}{\emph{IEEE Access}}  \bibinfo{volume}{10} (\bibinfo{year}{2022}), \bibinfo{pages}{16283--16298}.
\newblock


\bibitem[Yun et~al\mbox{.}(2019)]%
        {graphtransformer}
\bibfield{author}{\bibinfo{person}{Seongjun Yun}, \bibinfo{person}{Minbyul Jeong}, \bibinfo{person}{Raehyun Kim}, \bibinfo{person}{Jaewoo Kang}, {and} \bibinfo{person}{Hyunwoo~J Kim}.} \bibinfo{year}{2019}\natexlab{}.
\newblock \showarticletitle{Graph transformer networks}.
\newblock \bibinfo{journal}{\emph{Proceedings of NeurIPS}}  \bibinfo{volume}{32} (\bibinfo{year}{2019}).
\newblock


\bibitem[Zhang et~al\mbox{.}(2024a)]%
        {zhang2024mm}
\bibfield{author}{\bibinfo{person}{Chaoyi Zhang}, \bibinfo{person}{Kevin Lin}, \bibinfo{person}{Zhengyuan Yang}, \bibinfo{person}{Jianfeng Wang}, \bibinfo{person}{Linjie Li}, \bibinfo{person}{Chung-Ching Lin}, \bibinfo{person}{Zicheng Liu}, {and} \bibinfo{person}{Lijuan Wang}.} \bibinfo{year}{2024}\natexlab{a}.
\newblock \showarticletitle{Mm-narrator: Narrating long-form videos with multimodal in-context learning}. In \bibinfo{booktitle}{\emph{Proceedings of CVPR}}. \bibinfo{pages}{13647--13657}.
\newblock


\bibitem[Zhang et~al\mbox{.}(2023b)]%
        {zhang2023survey}
\bibfield{author}{\bibinfo{person}{Chaoning Zhang}, \bibinfo{person}{Fachrina~Dewi Puspitasari}, \bibinfo{person}{Sheng Zheng}, \bibinfo{person}{Chenghao Li}, \bibinfo{person}{Yu Qiao}, \bibinfo{person}{Taegoo Kang}, \bibinfo{person}{Xinru Shan}, \bibinfo{person}{Chenshuang Zhang}, \bibinfo{person}{Caiyan Qin}, \bibinfo{person}{Francois Rameau}, {et~al\mbox{.}}} \bibinfo{year}{2023}\natexlab{b}.
\newblock \showarticletitle{A survey on segment anything model (sam): Vision foundation model meets prompt engineering}.
\newblock \bibinfo{journal}{\emph{arXiv preprint arXiv:2306.06211}} (\bibinfo{year}{2023}).
\newblock


\bibitem[Zhang et~al\mbox{.}(2023a)]%
        {zhang2023video}
\bibfield{author}{\bibinfo{person}{Hang Zhang}, \bibinfo{person}{Xin Li}, {and} \bibinfo{person}{Lidong Bing}.} \bibinfo{year}{2023}\natexlab{a}.
\newblock \showarticletitle{Video-llama: An instruction-tuned audio-visual language model for video understanding}.
\newblock \bibinfo{journal}{\emph{arXiv preprint arXiv:2306.02858}} (\bibinfo{year}{2023}).
\newblock


\bibitem[Zhang et~al\mbox{.}(2024b)]%
        {zhang2024holmes}
\bibfield{author}{\bibinfo{person}{Huaxin Zhang}, \bibinfo{person}{Xiaohao Xu}, \bibinfo{person}{Xiang Wang}, \bibinfo{person}{Jialong Zuo}, \bibinfo{person}{Chuchu Han}, \bibinfo{person}{Xiaonan Huang}, \bibinfo{person}{Changxin Gao}, \bibinfo{person}{Yuehuan Wang}, {and} \bibinfo{person}{Nong Sang}.} \bibinfo{year}{2024}\natexlab{b}.
\newblock \showarticletitle{Holmes-VAD: Towards Unbiased and Explainable Video Anomaly Detection via Multi-modal LLM}.
\newblock \bibinfo{journal}{\emph{arXiv preprint arXiv:2406.12235}} (\bibinfo{year}{2024}).
\newblock


\bibitem[Zhao et~al\mbox{.}(2024)]%
        {jinyujiehawkeye}
\bibfield{author}{\bibinfo{person}{Jianing Zhao}, \bibinfo{person}{Jingjing Wang}, \bibinfo{person}{Yujie Jin}, \bibinfo{person}{Jiamin Luo}, {and} \bibinfo{person}{Guodong Zhou}.} \bibinfo{year}{2024}\natexlab{}.
\newblock \showarticletitle{Hawkeye: Discovering and Grounding Implicit Anomalous Sentiment in Recon-videos via Scene-enhanced Video Large Language Model}. In \bibinfo{booktitle}{\emph{Proceedings of ACM MM}}.
\newblock


\bibitem[Zhong et~al\mbox{.}(2023)]%
        {zhong2023attt2m}
\bibfield{author}{\bibinfo{person}{Chongyang Zhong}, \bibinfo{person}{Lei Hu}, \bibinfo{person}{Zihao Zhang}, {and} \bibinfo{person}{Shihong Xia}.} \bibinfo{year}{2023}\natexlab{}.
\newblock \showarticletitle{Attt2m: Text-driven human motion generation with multi-perspective attention mechanism}. In \bibinfo{booktitle}{\emph{Proceedings of the IEEE/CVF international conference on computer vision}}. \bibinfo{pages}{509--519}.
\newblock


\bibitem[Zhou et~al\mbox{.}(2023)]%
        {zhou2023dual}
\bibfield{author}{\bibinfo{person}{Hang Zhou}, \bibinfo{person}{Junqing Yu}, {and} \bibinfo{person}{Wei Yang}.} \bibinfo{year}{2023}\natexlab{}.
\newblock \showarticletitle{Dual memory units with uncertainty regulation for weakly supervised video anomaly detection}. In \bibinfo{booktitle}{\emph{Proceedings of AAAI}}, Vol.~\bibinfo{volume}{37}. \bibinfo{pages}{3769--3777}.
\newblock


\end{thebibliography}
